\documentclass{article} % For LaTeX2e
\usepackage{configuration/iclr2027_conference,times}

\usepackage{hyperref}
\usepackage{wrapfig}
\usepackage[utf8]{inputenc}
\usepackage{longtable}

\usepackage{amsmath,amssymb,amsthm}
\usepackage{bbm}
\newtheorem{theorem}{Theorem}[section]

\newtheorem{corollary}[theorem]{Corollary}

\newtheorem{proposition}[theorem]{Proposition}

\providecommand{\EEb}[2]{{\mathbb E}_{#1}\left[#2\right] } %

\providecommand{\PPb}[2]{{\rm Pr}_{#1}\left[#2\right] }

\providecommand{\hh}{\mathbf{h}}

\renewcommand{\ss}{\mathbf{s}}

\providecommand{\xx}{\mathbf{x}}
\providecommand{\yy}{\mathbf{y}}

\providecommand{\cL}{\mathcal{L}}

\newenvironment{talign*}
{\csname align*\endcsname}
{\endalign}

\usepackage[utf8]{inputenc}         %
\usepackage[T1]{fontenc}            %
\usepackage{url}                    %
\usepackage{booktabs}               %
\usepackage{amsfonts}               %
\usepackage{nicefrac}               %
\usepackage{microtype}              %
\usepackage{xcolor}                 %
\usepackage{algorithm}
\usepackage{algorithmic}
\usepackage{graphicx}
\usepackage{needspace}
\usepackage{subcaption}
\usepackage[flushleft]{threeparttable}
\usepackage{float}
\usepackage{multirow}
\usepackage{xspace}
\usepackage{natbib}
\usepackage{enumitem}
\usepackage[font=small]{caption}
\usepackage{autobreak}
\usepackage{sidecap}
\usepackage{wrapfig}
\usepackage{bbding}
\usepackage[toc, page, header]{appendix}
\usepackage{tikz}
\usepackage{xcolor}
\usepackage{colortbl}
\usepackage{pifont}
\usepackage{mdframed}

\hypersetup{
    colorlinks=true,
    linkcolor=blue,
    citecolor=blue,
    urlcolor=blue
}

\definecolor{coral}{RGB}{255,127,80}
\definecolor{darkgreen}{RGB}{0,100,0}
\definecolor{darkyellow}{RGB}{204,153,0}
\definecolor{salmon}{RGB}{250,128,114}
\definecolor{tableheader}{HTML}{EEF1F4}
\definecolor{tablegroup}{HTML}{E2E7EC}
\definecolor{methodlight}{HTML}{EDF4FB}
\definecolor{methodstrong}{HTML}{DCEAF7}
\definecolor{methodtext}{HTML}{245C8A}
\definecolor{gainaccent}{HTML}{B45F24}

\usepackage{xspace}

\usepackage[textwidth=3.5cm]{todonotes}

\newcommand{\method}{TPPO\xspace}

\title{TPPO: Tracking-Value PPO For Language Model Reasoning}

\author{%
  \bfseries Futing Wang\textsuperscript{1,2,3}\quad
  Chenxi Li\textsuperscript{3,4}\quad
  Jianhao Yan\textsuperscript{2,$\ddagger$}\quad
  Yun Luo\textsuperscript{3,$\dagger$,$\ddagger$}\quad
  Zhi Wang\textsuperscript{5}\quad
  Yizhuo Li\textsuperscript{3,6} \\
  \bfseries Runzhe Zhan\textsuperscript{7}\quad
  Yu Cheng\textsuperscript{7,$\ddagger$}\quad
  Tao Lin\textsuperscript{2,$\ddagger$} \\
  \normalfont Zhejiang University\textsuperscript{1}\quad
  Westlake University\textsuperscript{2}\quad
  Shanghai AI Laboratory\textsuperscript{3} \\
  \normalfont The Hong Kong University of Science and Technology\textsuperscript{4}\quad
  Nanjing University\textsuperscript{5} \\
  \normalfont Shanghai Jiao Tong University\textsuperscript{6}\quad
  Nanyang Technological University\textsuperscript{7}
}

\iclrfinalcopy % Display author information in the arXiv version.
\begin{document}

\begingroup
\setlength{\tabcolsep}{0pt} % Align the author block with the title's left edge.
\maketitle
\endgroup
\lhead{Preprint} % Mark the arXiv version as a preprint.
\begingroup
\renewcommand{\thefootnote}{\fnsymbol{footnote}}
\footnotetext[3]{\raggedright Corresponding authors.\quad\textsuperscript{$\dagger$}Project Leader.}
\endgroup

\begin{abstract}
% Proximal Policy Optimization (PPO) plays an important role in reinforcement learning with verifiable rewards for language-model reasoning.
% However, in long chain-of-thought tasks with sparse terminal rewards, its token-level critic can fail to track how the value changes as reasoning unfolds, leading
% to weak credit signals.
% In this work, we show that these changes have an additive form in log-odds space: each generated token contributes an update to the conditional
% probability of eventual success.
% Tracking-Value PPO (\method) uses this structure to learn token-level values through cumulative updates in log-odds space.
% \method uses the same terminal outcomes as PPO and requires no additional rollouts or process-level labels.
% Experiments on mathematical reasoning show that \method improves policy performance over PPO across model sizes and families.
% On Qwen3-4B-Base and Qwen3-8B-Base, it improves the aggregate score over PPO by 8.5 and 9.2 percentage points, respectively.
% Further analyses show improved value change capture and advantage quality during training.
Proximal Policy Optimization (PPO) is widely used in reinforcement learning with verifiable rewards for language-model reasoning.
While \emph{value evolution along trajectory} characterizes relative token-level credit, yet standard critics often fail to track it under sparse terminal rewards.
To model this evolution, we analyze the relation between successive prefix values and show that they admit a Bayesian decomposition with additive updates in log-odds space.
We use this additive structure to construct Tracking-Value PPO (\method), a cumulative value parameterization that sums position-wise scalar outputs into value logits.
These outputs are learned through terminal-outcome supervision on the reconstructed values.
% \method uses the same terminal outcomes as PPO, without additional rollouts or process-level supervision. 
% Across mathematical reasoning benchmarks, \method consistently outperforms PPO across model scales and variants, improving aggregate scores by 8.5 and 9.2 percentage points on Qwen3-4B-Base and Qwen3-8B-Base, respectively. These gains extend to countdown, science, and coding tasks. Further analysis confirms more accurate value-change prediction and advantage estimation.
\method improves aggregate mathematical reasoning scores over PPO by 7.0 and 8.4 points on Qwen3-4B-Base and Qwen3-8B-Base, respectively, while improving out-of-distribution generalization. Gains extend to Qwen3-4B, Qwen2.5-Math-1.5B, Qwen3.5-9B and to countdown, and scientific coding tasks.
Further analysis shows more accurate tracking of value evolution and more reliable advantage estimates.
Our code is publicly available in \url{https://github.com/LINs-lab/TPPO}.
\end{abstract}

\section{Introduction}

\label{sec:introduction}

Reinforcement Learning with Verifiable Rewards (RLVR) has substantially improved the reasoning capabilities of large language models (LLMs), enabling them to perform extended chain-of-thought (CoT) reasoning~\citep{guo2025deepseek, jaech2024openai}.
LLM reasoning often involves a long chain of token-level actions before arriving
at a solution, making credit assignment during reinforcement learning crucial.
% as in classical RL ~\citep{sutton1998reinforcement}.
In classical RL, the state-value function $V^{\pi}(\ss)$ provides a way to estimate the expected return from each state, thereby supporting credit assignment throughout the trajectory~\citep{sutton1998reinforcement}.
Proximal Policy Optimization (PPO)~\citep{schulman2017proximal} implements this idea with a critic network that estimates state values and uses them to compute advantages for policy updates.

\begin{figure}[t]
    \centering
    \includegraphics[width=1.0\linewidth]
{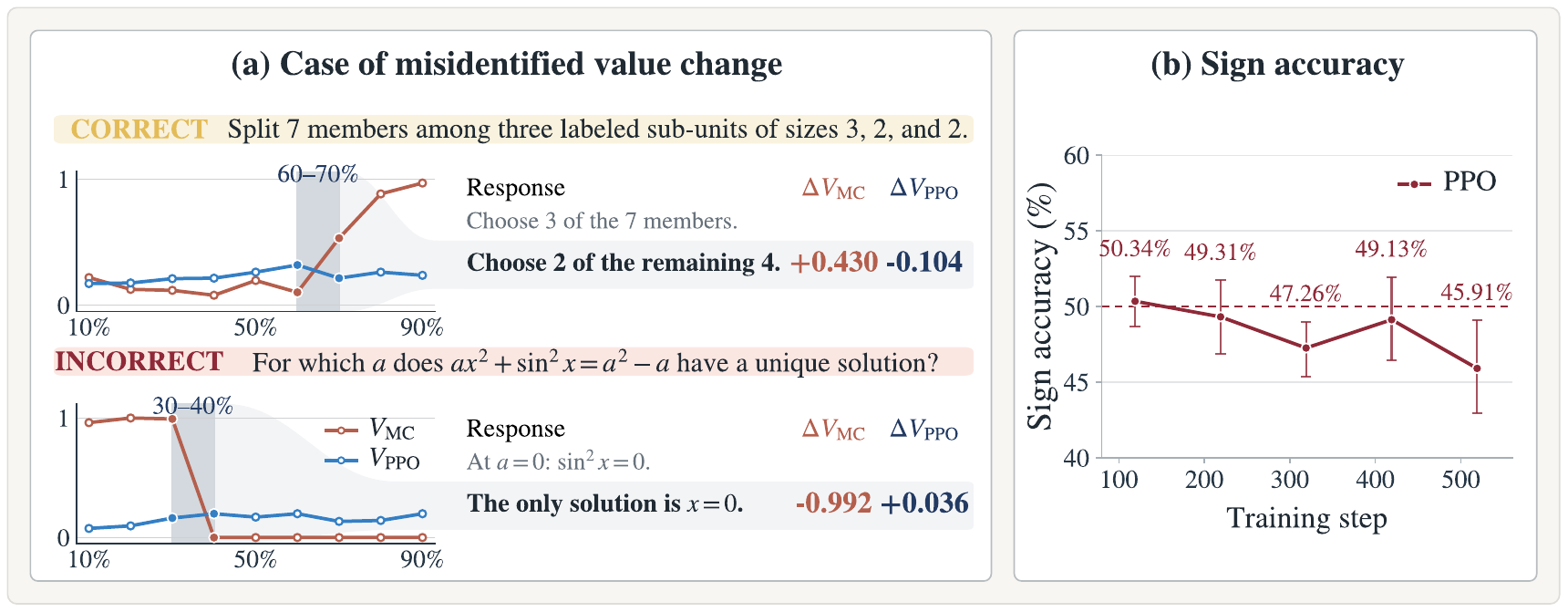}
        \caption{\small
           \textbf{Critics mispredict how value evolves along reasoning.}
           (a) Representative responses showing that different steps within a response can cause different MC-estimated value changes; the PPO critic predicts changes with opposite signs in these examples.
            (b) The sign accuracy of these predictions decreases during training.
            MC evaluation details are in Appendix~\ref{app:diagnostic_metrics}.
        }
        \label{fig:illustration}
        \vspace{-1.5em}
\end{figure}

% Credit assignment is a central problem in reinforcement learning and is particularly important for LLM reasoning, where a long sequence of token-level actions precedes the final answer.
% A standard approach to credit assignment in classical RL relies on learning a state-value function $V^{\pi}(s)$ to estimate the expected return from a given state~\citep{sutton1998reinforcement}.
% Proximal Policy Optimization (PPO)~\citep{schulman2017proximal} uses a critic network to estimate these values and compute advantages for policy updates.
However, assigning token-level credit with PPO remains challenging in long-CoT reasoning~\citep{yuan2025s, chen2026hista, wang2026sppo}.
More important than independently estimating each reasoning prefix, the key difficulty is tracking how value evolves along the reasoning trajectory, $\Delta V_t^\pi = V^\pi(\ss_{t+1}) - V^\pi(\ss_t)$, which characterizes token-level credit under policy $\pi$.
In standard PPO, the critic predicts each prefix value independently and does not explicitly model this evolution.
Empirically, Figure~\ref{fig:illustration}(a) illustrates a sharp drop from the correct step ``\textit{At $a=0$: $\sin^2 x=0$}'' to the incorrect claim ``\textit{The only solution is $x=0$}'', with $\Delta V_{\mathrm{MC}}=-0.992$. However, the PPO critic predicts $\Delta V_{\mathrm{PPO}}=+0.036$, missing the drop and reversing the MC-based GAE advantage ordering at the two anchors ($\gamma=\lambda=1$).
This problem is systematic: the sign accuracy remains near chance and decreases from $50.34\%$ to $45.91\%$ during training, as shown in Figure~\ref{fig:illustration}(b).
This observation motivates a central question:
\begin{center}
    \emph{How can a critic better model and learn value evolution across successive states?}
\end{center}

To answer this question, we characterize how state values evolve as reasoning unfolds (Section~\ref{sec:bayesian_strure_of_value}). Under binary terminal rewards, each prefix value is the probability that its continuation succeeds.
Conditioning on the next token updates this probability, and Bayes' rule shows that the resulting updates become additive when represented in log-odds space (Equation~\ref{eq:logit_value_update}).
To use this additive structure in modeling value evolution, we design \method (Tracking-Value PPO; Section~\ref{sec:method}), a critic parameterization that accumulates position-wise scalar outputs into value logits and applies a sigmoid to reconstruct prefix values, as illustrated in Figure~\ref{fig:absolute_vs_delta}.
The critic uses the same terminal outcomes as standard PPO and requires neither additional rollouts nor process-level annotations.
The reconstructed state values can be directly incorporated into GAE and the standard PPO objective.

% To utilize this structure, we propose \method (Tracking-Value PPO, Section~\ref{sec:method}).
% Instead of independently predicting the value at each reasoning state in standard critic parameterization, \method
% parameterizes the critic with token-wise logit changes and accumulates them to
% construct the value along the trajectory (illustrated in Figure~\ref{fig:absolute_vs_delta}).
% The critic is trained with the same terminal outcome supervision as standard
% PPO, \textit{without additional rollouts or process-level supervision}.
% The reconstructed values are then directly used for GAE and the standard PPO
% objective.

% Experiments across different models and reasoning tasks show that
% \method consistently improves policy performance over PPO baselines (Section~\ref{sec:experiments}).
% Further analysis shows that \method better captures value changes and produces
% more accurate advantage estimates.
Extensive experiments with Qwen3-4B-Base, Qwen3-8B-Base, and Qwen3-4B demonstrate the effectiveness of \method across diverse reasoning tasks (Section~\ref{sec:experiments}).
On mathematical reasoning, \method improves both in-distribution performance and out-of-distribution generalization.
These gains extend consistently to countdown, scientific coding tasks, where \method outperforms the corresponding PPO baselines.
Further analysis shows that \method more accurately tracks how values evolve along reasoning, leading to more reliable advantage estimates.
Our contributions are as follows:
\looseness=-1

\begin{itemize}[leftmargin=*,nosep]

\item \textbf{Characterizing how value evolves along reasoning.}
We reveal an empirical discrepancy between Monte Carlo estimates of value evolution
and the evolution captured by standard PPO critics, highlighting the challenge of
learning fine-grained credit signals under sparse rewards.

\item \textbf{A cumulative parameterization of prefix values.}
We use the additive structure of prefix values in log-odds space to construct
\method, a critic that directly modeling value evolution and accumulates learned scalar increments and is trained with terminal-outcome supervision.

\item \textbf{Improved policy optimization.}
Experiments across mathematical reasoning, Countdown, and scientific coding demonstrate that \method improves reasoning performance and enhances the quality of training
signals across models.

\end{itemize}

\section{Related Work}
\label{sec:related_work}

\paragraph{Reinforcement Learning and Credit Assignment for LLM Reasoning}
% rlvr category
Reinforcement learning has become a widely used approach for improving LLM reasoning.  Value-free methods such as GRPO~\citep{shao2024deepseekmath} and DAPO~\citep{yu2026dapo} replace learned value with intra-group reward.
They treat the entire response as a single action and apply the same state value to every token.
Value-based methods such as Proximal Policy Optimization~\citep{schulman2017proximal} use a learned value function to estimate advantages via GAE~\citep{schulman2015high}.
Under sparse terminal rewards, assigning credit to intermediate states remains challenging.
One line of work uses Monte Carlo sampling or tree search to obtain more accurate value estimates for intermediate states~\citep{kazemnejad2024vineppo, hou2506treerl, li2025treepo}.
These methods require additional sampling or search during training.
Another line of work introduces process-level supervision to provide denser feedback over intermediate reasoning steps~\citep{lightman2024let, wang2024math, zhang2025lessons, setlur2025rewarding, qu2025optimizing, cui2025process, chen2026hista}.
Such methods typically rely on additional process supervision or auxiliary reward models.
% In contrast, our work improves credit assignment through value-function learning itself, without additional supervision or rollout computation.
In contrast, we focus on value-function learning as a way to improve credit assignment, without introducing additional supervision or rollout computation.

\paragraph{Value Function Learning}
Prior work also studied how value function should be learned.
Distributional RL~\citep{bellemare2017distributional} models distributions over returns, while Stop Regressing~\citep{farebrother2024stop} shows that categorical classification can improve value learning. HL-Gauss PPO~\citep{zhou2026start} recently applies classification-based value learning to PPO critics for LLM reinforcement learning. VAPO~\citep{yue2025vapo} develops a value-based PPO recipe for long-chain-of-thought training, addressing value bias, heterogeneous sequence lengths, and sparse rewards through its training design.
SPPO~\citep{wang2026sppo} reformulates long-horizon reasoning as a
sequence-level contextual bandit and uses a decoupled scalar value function
to avoid unstable token-level credit assignment in PPO.
\method addresses a different aspect of value learning: it parameterizes prefix values through cumulative scalar outputs in log-odds space.

\section{PPO Critics Struggle to Track Value Evolution}
\label{sec:characterizing_failure}
% In this section, we introduce the value function and its changes during reasoning, review how PPO learns values from sampled returns, and figure out an empirical mismatch between Monte Carlo value change and the changes captured by the learned critic.
In this section, we formulate autoregressive reasoning as a token-level Markov
decision process and review PPO value estimation for token-level credit assignment.
For distinguishing credit between successive tokens, we focus on value evolution across states.
Standard PPO critics do not explicitly model this evolution, and our empirical analysis shows that they fail to track it.
\looseness=-1
\subsection{Values and Token-level Credit}
\label{sec:value_and_credit}
% \paragraph{Value as success probability.}
We model autoregressive generation as a token-level Markov decision process~\citep{trung2024reft, jaech2024openai}.
Given a prompt $\xx$, the policy $\pi_\theta$ samples a response
$\yy=(y_1,\ldots,y_T)$ for a prompt $\xx$.
At time $t$, the state is $\ss_t=(\xx,\yy_{<t})$ and the token action is
$a_t=y_t\sim\pi_\theta(\cdot\mid\ss_t)$.
In the reinforcement learning with verifiable rewards setting, a scalar reward is assigned only at the terminal step: $r_t=0$ for $t<T$ and $r_T=R(\xx, \yy)\in\{0,1\}$, with the discount factor set to $\gamma=1$~\citep{guo2025deepseek,shao2024deepseekmath, ouyang2022training}.
\looseness=-1
\paragraph{Policy Gradient and Advantage Functions.}
Given the MDP formulation, policy gradient methods optimize the expected terminal reward by assigning a credit signal to each sampled token:
\begin{align}
    J(\theta)
    &\doteq
    \EEb{\xx\sim\mathcal{D},\,\yy\sim\pi_\theta(\cdot\mid\xx)}{R(\xx, \yy)},
    \label{eq:policy_objective}\\
    \mathbf{g}_{\mathrm{pg}}
    \doteq
    \nabla_\theta J(\theta)
    &=
    \EEb{\xx\sim\mathcal{D},\,\yy\sim\pi_\theta(\cdot\mid\xx)}{
        \sum_{t=1}^{T}
        \nabla_\theta\log\pi_\theta(a_t\mid\ss_t)
        A^{\pi_\theta}(\ss_t,a_t)
    }.
    \label{eq:policy_gradient}
\end{align}
\looseness=-1
Here, $\mathcal{D}$ is the prompt distribution, and the advantage function $A^{\pi_\theta}(\ss_t,a_t)$ measures how much better token $a_t$ is than the policy's average choice at state $\ss_t$.
% Formally,
% \begin{equation}
%     A^\pi(\ss_t,a_t)\doteq Q^\pi(\ss_t,a_t)-V^\pi(\ss_t),
%     \label{eq:advantage_definition}
% \end{equation}
% where $Q^\pi(\ss_t,a_t)$ is the expected return after choosing $a_t$ and then following $\pi$, and $V^\pi(\ss_t)$ is the expected return under $\pi$ from $\ss_t$.
The sign and absolute magnitude of $A^{\pi_\theta}(\ss_t,a_t)$ determine the direction and weight of the token's contribution to the policy gradient.
\vspace{-0.5em}

\paragraph{Advantage Estimation and Value Functions.}
Generalized Advantage Estimation (GAE)~\citep{schulman2015high} is a widely used advantage estimator that controls the bias--variance trade-off by combining temporal-difference (TD) residuals.
With critic predictions $\widehat V_t=V_{\phi}(\ss_t)$, $\gamma=\lambda=1$,
and terminal rewards, the GAE estimate reduces to:
\begin{equation}
    \begin{aligned}
    \widehat A_t^{\mathrm{GAE}}
    &=\sum_{\ell=0}^{T-t}(\gamma\lambda)^\ell\widehat\delta_{t+\ell}
    &=\sum_{k=t}^{T}\widehat\delta_k=R(\xx,\yy)-\widehat V_t.
    \end{aligned}
    \label{eq:gae_estimator}
\end{equation}
where $\widehat\delta_t\doteq r_t+\gamma\widehat V_{t+1}-\widehat V_t$ is the critic-based one-step TD residual, with $\widehat V_{T+1}=0$ at the terminal state.
% Smaller $\lambda$ typically reduces sampling variance, while larger $\lambda$ reduces reliance on intermediate value estimates and the bias they may introduce.
The true value $V^{\pi_\theta}(\ss_t)$ is the expected return from the current prefix.
In this setting, the return is $G_t\doteq\sum_{k=t}^{T}r_k=R(\xx,\yy)$, so
\begin{equation}
    V^{\pi_\theta}(\ss_t)
    \doteq\EEb{\pi_\theta}{G_t\mid S_t=\ss_t}
    =\PPb{\pi_\theta}{R(\xx,\yy)=1\mid S_t=\ss_t}.
    \label{eq:value_as_probability}
\end{equation}
Thus, $V^{\pi_\theta}(\ss_t)$ is the probability that a continuation from the current prefix succeeds.
\vspace{-0.5em}

\paragraph{Estimating Advantage via Value Networks.}
Since the true value $V^{\pi_\theta}(\ss_t)$ is unavailable in practice, PPO learns a value network $V_{\phi}$ to approximate $V^{\pi_\theta}(\ss_t)$.
Its predictions are used in Equation~\ref{eq:gae_estimator} to obtain advantage estimates.
The network parameters $\phi$ are optimized alongside the policy $\pi_\theta$ by minimizing the mean squared error between predicted values and sampled returns $G_t$.
For a batch $\mathcal{D}_{\pi_\theta}=\{\yy^{(n)}\}_{n=1}^N$, with response $n$ having $T_n$ valid response tokens, this objective is
\begin{equation}
    \widehat{\cL}^{\mathrm{VF}}(\phi)
    \doteq
    \frac{1}{\sum_{n=1}^N T_n}
    \sum_{n=1}^N\sum_{t=1}^{T_n}
    \left(
        V_{\phi}(\ss_t^{(n)})
        -G_t^{(n)}
    \right)^2.
    \label{eq:mc_value_loss}
\end{equation}
Under terminal rewards, $G_1^{(n)}=\cdots=G_{T_n}^{(n)}=R(\yy^{(n)})$, every state visited by the same response receives the same observed binary target.
Sharing a sampled return does not imply equal values: the expected target at each prefix remains $V^{\pi_\theta}(\ss_t)$.
\looseness=-1
% This regression does not explicitly model how the values of successive states are related.

\subsection{PPO Fails to Track Value Evolution}
\paragraph{Value evolution determines relative token-level credit.}
Although all tokens in a response share the same terminal reward, they need not deserve the same credit.
Value evolution is the change in the probability of eventual success as each
token extends the prefix.
Different steps in the same response can therefore have very different effects: Figure~\ref{fig:illustration}(a) shows a sharp
MC-estimated drop across the incorrect claim ``The only solution is $x=0$''
($\Delta V_{\mathrm{MC}}=-0.992$), but an increase across the correct step
``Choose 2 of the remaining 4'' ($\Delta V_{\mathrm{MC}}=+0.430$).
These changes indicate which steps should receive less or more relative credit.
From Equation~\ref{eq:gae_estimator}, the change equals the difference between
adjacent tokens' advantages: for $t<T$,
\begin{equation}
    \widehat A_t^{\mathrm{GAE}}-\widehat A_{t+1}^{\mathrm{GAE}}
    =\widehat V_{t+1}-\widehat V_t
    \doteq\Delta\widehat V_t.
    \label{eq:gae_value_change}
\end{equation}
$\Delta\widehat V_t>0$ indicates increased success probability and gives $a_t$
a larger estimated advantage than $a_{t+1}$; $\Delta\widehat V_t<0$ indicates a
decrease and reverses this ordering.
% Predicting the opposite sign therefore reverses their relative credit.
Thus, predicting the opposite sign of $\Delta\widehat V_t$ in Figure~\ref{fig:illustration} indicates a reversal of the relative credit.
Since advantages weight token gradients in Equation~\ref{eq:policy_gradient},
this misordering changes the relative weights of their contributions to the
sampled policy update.
We discuss in Appendix~\ref{app:value_change_credit} how value evolution affects advantage estimation for different $\lambda$.
\vspace{-1em}

% \begin{remark}[What the PPO Critic Learns]
%     At every visited prefix, PPO regresses to the sampled response's terminal outcome $G_t=R(\yy)$; across responses, the population target is
%     $V^\pi(\ss)=\EEb{\pi}{G_t\mid S_t=\ss}$, the success probability of a continuation under $\pi$.
%     Equation~\ref{eq:mc_value_loss} is the resulting batch regression, with one Bernoulli return sample at each visited prefix.
% \end{remark}
\paragraph{Empirical Observation}
We evaluate whether PPO critics recover the direction of local value evolution.
We estimate prefix values using Monte Carlo (MC) continuations~\citep{kazemnejad2024vineppo,chen2026hista} at fixed
fractions of each response and measure sign agreement between MC-estimated
and critic-predicted differences across adjacent anchors.
Appendix~\ref{app:coarse_value_change} details the estimation procedure and
the connection between anchor-level changes and relative GAE advantages.
Using this protocol, we find that the sign errors shown in
Figure~\ref{fig:illustration}(a) persist across training: aggregate sign
accuracy stays near chance, falling from
$50.34\%$ to $45.91\%$ between the first and last evaluated checkpoints
(Figure~\ref{fig:illustration}(b)).
Continued training does not reliably recover the direction of value evolution, and poor value training introduces noisy credit assignment.
This suggests that value evolution itself should be explicitly captured in the critic.
\vspace{-0.5em}
% This motivates explicitly modeling the relation between successive prefix values.
% We therefore ask:
% \textbf{Can explicitly modeling the relation between successive values
% improve the critic's tracking of value evolution?}

% \begin{figure}[H]
%     \centering
%     \includegraphics[width=1.0\linewidth]
%     {sources/figures/outputs/01_ppo_critic_failure_and_outcomes_train_seed234.pdf}
%     \caption{\small
%         \textbf{The same critic failure on the training-seed evaluation.}
%         The training-seed split reproduces the mismatch between the nearly constant
%         PPO critic and the outcome-conditioned Monte Carlo value trajectory.
%     }
%     \label{fig:ppo_failure_train_seed}
% \end{figure}

\section{\method: Tracking-Value PPO}
\label{sec:deltav_method}
Our goal is to explicitly model value evolution across successive prefixes.
Analyzing the relation between $V^{\pi_\theta}(\ss_t)$ and
$V^{\pi_\theta}(\ss_{t+1})$, we find a Bayesian update structure that is
additive in log-odds space. We use this structure to construct the cumulative
logit parameterization of \method.
\vspace{-0.5em}
\subsection{Bayesian Structure of Value Evolution}
\label{sec:bayesian_strure_of_value}
Recall from Equation~\ref{eq:value_as_probability} that
$V^{\pi_\theta}(\ss_t)$ is the conditional probability that a continuation from
$\ss_t$ eventually succeeds.
After sampling token $a_t$, the next state
$\ss_{t+1}=(\ss_t,a_t)$ conditions this probability on one additional
observation.
The transition from $V^{\pi_\theta}(\ss_t)$ to $V^{\pi_\theta}(\ss_{t+1})$ can therefore
be written as a Bayesian update.

\begin{proposition}[Bayesian update of value]
\label{prop:logit_value_update}
Assume a fixed policy $\pi_\theta$ (with $\theta$ held fixed), a binary terminal
reward, $\gamma=1$, and $0<V^{\pi_\theta}(\ss_t)<1$. For any sampled token
$a_t$ for which the conditional
probabilities below are positive,
\begin{equation}
    V^{\pi_\theta}(\ss_{t+1})
    =
    \frac{
        V^{\pi_\theta}(\ss_t)\PPb{\pi_\theta}{a_t\mid\ss_t,R=1}
    }{
        V^{\pi_\theta}(\ss_t)\PPb{\pi_\theta}{a_t\mid\ss_t,R=1}
        +
        \bigl(1-V^{\pi_\theta}(\ss_t)\bigr)
        \PPb{\pi_\theta}{a_t\mid\ss_t,R=0}
    }.
    \label{eq:bayesian_value_update}
\end{equation}
Equivalently, taking log-odds yields
\begin{equation}
    \operatorname{logit}V^{\pi_\theta}(\ss_{t+1})
    -\operatorname{logit}V^{\pi_\theta}(\ss_t)
    =
    \log
    \frac{\PPb{\pi_\theta}{a_t\mid\ss_t,R=1}}
         {\PPb{\pi_\theta}{a_t\mid\ss_t,R=0}}
    \doteq \Delta_{t+1}^{\pi_\theta}.
    \label{eq:logit_value_update}
\end{equation}
\end{proposition}
The derivation, together with the boundary cases
$V^{\pi_\theta}(\ss_t)\in\{0,1\}$, is provided in
Appendix~\ref{app:bayesian_value_update}.
Equation~\ref{eq:logit_value_update} characterizes the true values under a
fixed policy: their successive log-odds changes equal token-specific
log-likelihood ratios.
Applying this identity recursively along the trajectory yields the following
decomposition.
\looseness=-1

\subsection{Method}
\label{sec:method}

To utilize the structure of value evolution derived in
Proposition~\ref{prop:logit_value_update}, we first unroll the Bayesian
updates along a response.

\begin{corollary}[Incremental decomposition of values]
\label{cor:value_decomposition}
Applying Equation~\ref{eq:logit_value_update} recursively gives
\begin{equation}
    V^{\pi_\theta}(\ss_t)
    =
    \sigma\!\left(
        \sum_{i=1}^{t} c_i^{\pi_\theta}
    \right), \quad
    c_1^{\pi_\theta}=\operatorname{logit}V^{\pi_\theta}(\ss_1),
    \quad
    c_i^{\pi_\theta}=\Delta_i^{\pi_\theta},\quad i>1 .
\end{equation}
\end{corollary}

\begin{figure}[t]
    \centering
    \includegraphics[width=0.95\linewidth]
    {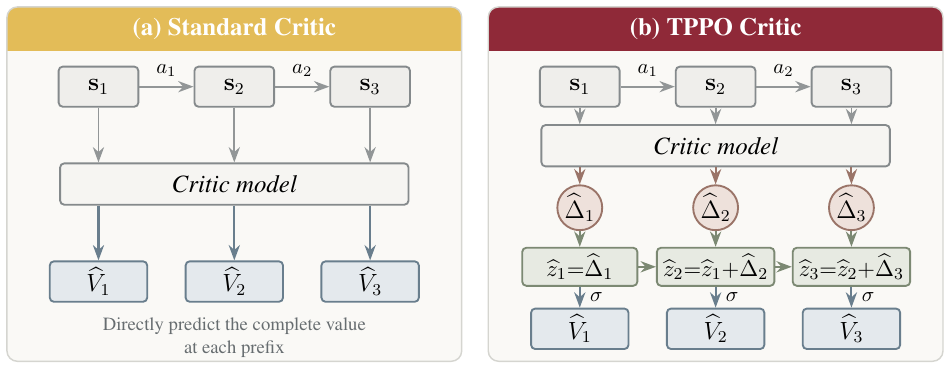}
    \caption{\small
        \textbf{Difference between standard and \method critic parameterizations.}
        (a) A standard critic directly predicts a value for each prefix.
        (b) \method accumulates token-wise critic outputs into value logits,
        then applies a sigmoid to obtain prefix values.
    }
    \label{fig:absolute_vs_delta}
    \vspace{-1.2em}
\end{figure}

To model value evolution using this cumulative form, we propose Tracking-Value PPO (\method), which parameterizes the critic as an accumulation of position-wise scalar outputs: $\widehat V_t = \sigma\!\left( \sum_{i=1}^{t} f_\phi(\ss_i) \right)$.
The resulting increments describe the model's own value sequence; the
Bayesian identity does not guarantee that learned outputs recover the true
log-likelihood ratios in Equation~\ref{eq:logit_value_update}.
The comparison with a standard critic model is illustrated in Figure~\ref{fig:absolute_vs_delta}.

% \method makes token-wise value changes explicit in the critic
% parameterization, and learns them implicitly through value reconstruction.
% No additional targets for the value changes are required.

% \paragraph{Critic Loss.}
Under binary terminal rewards,
$V^{\pi_\theta}(\ss_t)=\Pr_{\pi_\theta}(R=1\mid\ss_t)$.
We train each reconstructed value against the terminal outcome
$R(\yy)\in\{0,1\}$ using
\begin{equation}
    \mathcal{L}_{V}
    =
    -\frac{1}{|\mathcal{T}|}
    \sum_{t\in\mathcal{T}}
    \left[
        R(\yy)\log \widehat V_t
        +
        \bigl(1-R(\yy)\bigr)
        \log\bigl(1-\widehat V_t\bigr)
    \right],
    \label{eq:deltav_loss}
\end{equation}
where $\mathcal{T}$ denotes the valid response positions.
% In practice, we evaluate the equivalent objective in logit space using
% $\operatorname{BCEWithLogits}$.
The scalar outputs are learned jointly through this value objective, without
explicit targets for log-odds changes. Each increment receives gradients
from all subsequent reconstructed values that depend on it
(Appendix~\ref{app:implicit_delta_learning}).
\vspace{-0.5em}

\paragraph{Cumulative parameterization improves the tracking of value evolution.}
To isolate the effect of parameterization, we fix the actor and train critics using either the standard parameterization with
MSE or BCE, or the cumulative parameterization (\method) with BCE.
For the BCE objective, the standard critic uses a sigmoid output to match
the bounded value targets.
We conduct this controlled comparison on the Qwen3-8B-Base mathematical
reasoning task.
As shown in Table~\ref{tab:fixed_actor_critic}, \method improves the directional accuracy of local value evolution over the standard BCE critic by 2.01 and 2.80 percentage points at $\tau=0$ and $\tau=1$, respectively.
Because both critics use the same BCE objective, this improvement isolates the contribution of the cumulative parameterization and shows that its benefit extends beyond the choice of loss function.
\vspace{-0.5em}
\begin{table}[t]
    \centering
    \caption{\textbf{Directional accuracy of local value evolution under a fixed actor.}
    Entries are accuracies (\%) with 95\% confidence intervals in brackets,
    computed on intervals satisfying $\mathrm{SNR}>\tau$.
    The metric compares the signs of critic-predicted and MC-estimated
    local value evolution between evaluation anchors
    (Appendix~\ref{app:diagnostic_metrics}).}
    \label{tab:fixed_actor_critic}
    \small
    \setlength{\tabcolsep}{10pt}
    \renewcommand{\arraystretch}{1.15}
    \begin{tabular}{@{}lcc@{}}
        \toprule
        Critic + loss & Sign accuracy ($\tau=0$) & Sign accuracy ($\tau=1$) \\
        \midrule
        Standard critic + MSE & 50.51 [49.15, 51.79] & 51.19 [49.57, 52.69] \\
        Standard critic + BCE & 50.70 [49.29, 52.13] & 51.14 [49.54, 52.74] \\
        \method critic + BCE & \textbf{52.71} [51.67, 53.77] & \textbf{53.94} [52.59, 55.33] \\
        \bottomrule
    \end{tabular}
    \vspace{-1.3em}
\end{table}

\paragraph{Train with PPO}
The resulting $\widehat V_t$ is used in the standard GAE estimator, and the
policy is optimized with the usual PPO clipped objective~\citep{schulman2017proximal},
\begin{equation}
    \mathcal{L}_{\mathrm{PPO}}
    =
    -\frac{1}{|\mathcal{T}|}
    \sum_{t\in\mathcal{T}}
    \min\!\left(
        \rho_t(\theta)\widehat A_t,\,
        \operatorname{clip}
        \bigl(\rho_t(\theta),1-\epsilon,1+\epsilon\bigr)\widehat A_t
    \right),
    \label{eq:tppo_policy_loss}
\end{equation}
where
$\rho_t(\theta)= \pi_\theta(a_t\mid\ss_t)/ \pi_{\mathrm{old}}(a_t\mid\ss_t)$.
Apart from the critic parameterization, its advantage estimation and policy optimization follow standard PPO.

\section{Experiments}
\label{sec:experiments}
\vspace{-0.5em}
In this section, we evaluate \method in terms of reasoning performance,
the critic's ability to track value evolution, and the quality of
advantage estimates.
Specifically, we ask:
\begin{itemize}[leftmargin=*,nosep]
\item \textbf{RQ1: Policy performance.}
Does \method improve policy performance across model scales and tasks?

\item \textbf{RQ2: Tracking value evolution.}
Does the cumulative critic better track value evolution and estimate
prefix values along reasoning trajectories?

\item \textbf{RQ3: Credit assignment.}
Do better value estimates and value-evolution tracking produce more accurate
advantage estimates for policy updates?
\end{itemize}
\vspace{-0.5em}

\subsection{Experimental Setup}
\label{sec:experimental_setup}
\vspace{-0.2em}
\paragraph{Evaluation.}
\vspace{-0.3em}
For mathematical reasoning, we evaluate on ten benchmarks: in-domain tasks
(AIME24/25/26, AMC~\citep{li2024numinamath}, MATH-500~\citep{hendrycks2021measuring},
Minerva~\citep{lewkowycz2022solving}, and OlympiadBench~\citep{he2024olympiadbench})
and OOD tasks (ARC-C~\citep{clark2018think}, AGI-Eval~\citep{zhong2024agieval},
and BBH~\citep{suzgun2023challenging}). We report Avg@32 on AIME and AMC,
Avg@4 on the remaining benchmarks, using temperature 1.0 and a maximum
response length of 24k tokens.
For scientific coding, we evaluate SciCode~\citep{SciCode}, HumanEval~\citep{humaneval},
HumanEvalFix~\citep{HumanEvalFix}, LiveCodeBench v6~\citep{LiveCodeBench},
and APPS Introductory~\citep{APPS}, averaging three evaluation runs.

\paragraph{RLVR setup.}
For mathematical reasoning, we train on approximately 30k examples from Polaris~\citep{Polaris2025},
with $n\in\{1,8\}$ sampled responses per prompt.
Training responses are sampled at temperature 1.0 with a maximum length
of 8,192 tokens. We use Math-Verify\footnote{\url{https://github.com/huggingface/Math-Verify}}
to compute outcome rewards for mathematical reasoning.
Experiments use the Slime framework~\citep{slime_github} on nodes with eight H200 GPUs each.
For scientific coding, we train on data from Sciwalker~\citep{li2026sciwalkersynthesizingscientificcoding}.
Further implementation details are provided in Appendix~\ref{app:experimental_details}.
In all main experiments, the position-wise outputs use a shared scalar head.
We discuss and evaluate a separate-head alternative in Appendix~\ref{app:shared_head}.

\paragraph{Models and baselines.}
Our main experiments use Qwen3-4B-Base and Qwen3-8B-Base~\citep{yang2025qwen3}.
The $n=1$ setting also includes Qwen3-4B, with thinking disabled during training.
We compare \method with PPO~\citep{schulman2017proximal},
GRPO~\citep{shao2024deepseekmath}, SPPO~\citep{wang2026sppo}, and
VAPO~\citep{yue2025vapo}, using the same training dataset and evaluation protocol
within each setting. We also report each model's performance before RL
training.
To test \method on another task and a smaller model, we train
Qwen2.5-Math-1.5B on Countdown problems with three or four input numbers
and evaluate on problems with three, four, or five input numbers.
% Confirm the public Polaris dataset release and the run-specific training budgets.

\paragraph{Critic evaluation.}
At five Qwen3-8B-Base training checkpoints, we evaluate PPO and \method on
a fixed subset of 64 prompts from the training set.
The same prompts are used for both methods at every checkpoint, with each method's actor sampling 8 responses per prompt, yielding approximately 5.1k rollouts
across both methods and all five checkpoints.
We estimate MC values at nine anchors spanning 10--90\% of each response, allocating 128 continuations per anchor (approximately 5.9M valid MC continuations in total).
We measure agreement with these estimates in the direction of local value evolution and in advantages.
Sampling details, metric definitions, and bootstrap estimation of the gains over PPO are provided in Appendix~\ref{app:diagnostic_metrics}.

\providecommand{\ppoGain}[1]{\textcolor{gainaccent}{\scriptsize\,\textbf{(#1)}}}
\begin{table}[t]
    \centering
    \caption{
        \textbf{In-domain and out-of-domain results with $n=8$.}
        All entries are accuracies (\%).
        Orange parenthetical values show the absolute improvement over PPO
        in percentage points.
        Bold and underlined values indicate the best and second-best
        results within each model group, respectively.
    }
    \label{tab:main_results}
    \small
    \setlength{\tabcolsep}{2.45pt}
    \renewcommand{\arraystretch}{1.25}
    \resizebox{\linewidth}{!}{%
    \begin{tabular}{@{}lcccccc|cccc@{}}
        \toprule
        \rowcolor{tableheader}
        \textbf{Method}
        & \multicolumn{6}{c|}{\textbf{In-Domain Performance}}
        & \multicolumn{4}{c}{\textbf{Out-of-Domain Performance}} \\
        \rowcolor{tableheader}
        & \textbf{AIME24/25/26}
        & \textbf{AMC}
        & \textbf{MATH500}
        & \textbf{Minerva}
        & \textbf{Olympiad}
        & \textbf{Avg.}
        & \textbf{ARC-C}
        & \textbf{AGI-Eval}
        & \textbf{BBH}
        & \textbf{OOD Avg.} \\
        \midrule
        \rowcolor{tablegroup}
        \multicolumn{11}{c}{\textbf{Qwen3-4B-Base}} \\
        \midrule
        Base
        & 4.3/3.3/3.4 & 27.3 & 43.2
        & 22.1 & 22.3 & 18.0
        & 34.8 & 24.8 & 12.8 & 24.1 \\
        GRPO
        & 12.1/11.4/8.1 & 51.9 & 76.9
        & \textbf{47.6} & 40.1 & 35.4
        & \underline{90.2} & \underline{63.7} & \underline{42.9} & \underline{65.6} \\
        PPO
        & \underline{13.5}/9.5/\underline{9.3} & 53.8 & 76.1
        & 41.2 & 39.1 & 34.6
        & 88.7 & 59.0 & 40.7 & 62.8 \\
        VAPO
        & 8.3/8.3/6.7 & 48.8 & 72.2
        & 44.9 & 34.9 & 32.0
        & 88.7 & 58.1 & 41.2 & 62.7 \\
        SPPO
        & 9.2/\underline{15.0}/7.5 & \underline{58.1} & \underline{77.5}
        & 40.8 & \underline{41.3} & \underline{35.6}
        & 90.1 & 63.0 & 40.5 & 64.5 \\
        \rowcolor{methodstrong}
        \textcolor{methodtext}{\textbf{\method}}
        & \textbf{18.5}/\textbf{18.4}/\textbf{18.5}
        & \textbf{62.5} & \textbf{81.8}
        & \underline{46.5} & \textbf{45.3}
        & \textbf{41.6}\,\ppoGain{+7.0}
        & \textbf{90.6} & \textbf{66.5}
        & \textbf{43.7} & \textbf{66.9} \\
        % \rowcolor{methodstrong}
        % \textcolor{methodtext}{\textbf{\method + PI}}
        % & \textbf{25.5}/\textbf{20.4}/\underline{15.9}
        % & \textbf{63.1} & \textbf{83.0}
        % & \underline{46.6} & \textbf{47.2}
        % & \textbf{43.1}\,\ppoGain{+8.5}
        % & \underline{58.5} & \underline{65.0}
        % & \underline{42.5} & \underline{55.3} \\
        \midrule
        \rowcolor{tablegroup}
        \multicolumn{11}{c}{\textbf{Qwen3-8B-Base}} \\
        \midrule
        Base
        & 8.2/6.8/6.4 & 33.8 & 54.3
        & 24.9 & 27.2 & 23.1
        & 62.4 & 44.2 & 27.8 & 44.8 \\
        GRPO
        & \underline{19.7}/14.1/9.4 & 59.5 & 79.6
        & 46.0 & 41.2 & \underline{38.5}
        & \underline{93.0} & 67.7 & 34.8 & 65.2 \\
        PPO
        & 15.8/12.5/\underline{11.3} & \underline{61.6} & 78.8
        & 40.8 & 42.0 & 37.5
        & 92.2 & 68.4 & \underline{42.2} & 67.6 \\
        VAPO
        & 14.2/13.3/8.3 & 59.4 & 79.6
        & \underline{47.4} & 40.8 & 37.6
        & 91.9 & 67.6 & 40.6 & 66.7 \\
        SPPO
        & 14.2/\underline{14.2}/7.5 & 61.3 & \underline{80.1}
        & 45.6 & \underline{42.8} & 38.0
        & 92.9 & \underline{69.9} & 40.7 & \underline{67.8} \\
        \rowcolor{methodstrong}
        \textcolor{methodtext}{\textbf{\method}}
        & \textbf{25.8}/\textbf{22.3}/\textbf{20.5}
        & \textbf{67.3} & \textbf{85.4}
        & \textbf{49.8} & \textbf{50.2}
        & \textbf{45.9}\,\ppoGain{+8.4}
        & \textbf{93.1} & \textbf{72.0}
        & \textbf{43.9} & \textbf{69.7} \\
        \rowcolor{methodstrong}
        % \textcolor{methodtext}{\textbf{\method + PI}}
        % & \textbf{28.3}/\underline{19.9}/\textbf{21.3}
        % & \textbf{69.8} & \textbf{86.2}
        % & \underline{48.4} & \textbf{52.8}
        % & \textbf{46.7}\,\ppoGain{+9.2}
        % & \textbf{64.3} & \textbf{72.9}
        % & \textbf{44.2} & \textbf{60.5} \\
        \bottomrule
    \end{tabular}%
    \vspace{-6em}
    }
\end{table}

\begin{figure}[t]
    \centering
    
    \includegraphics[width=1.0\linewidth]{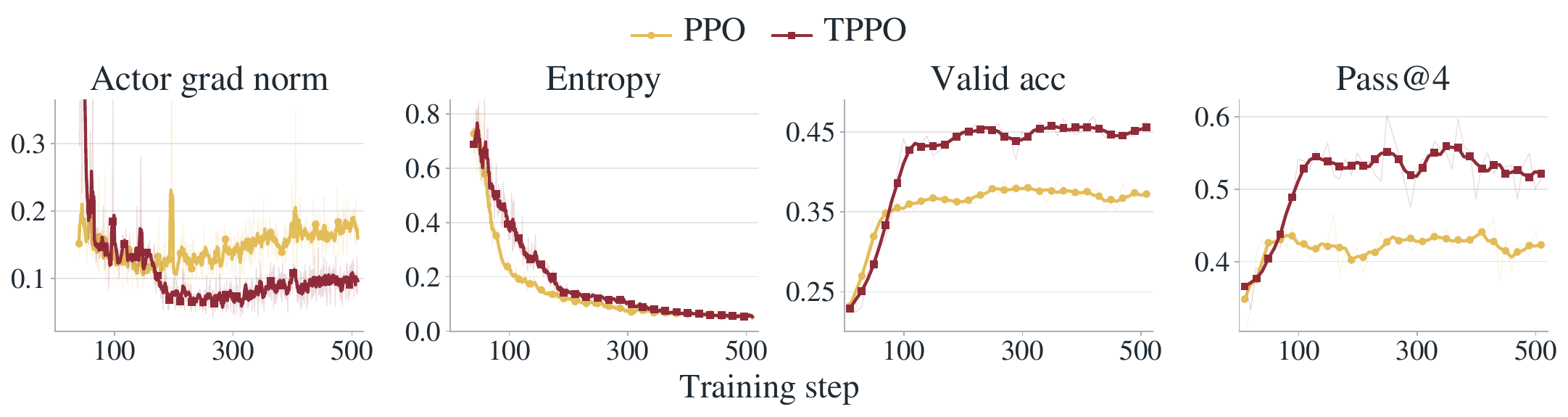}
    \caption{\small
        \textbf{Training dynamics comparing PPO and \method on Qwen3-8B-Base ($n=8$).}
        We track actor gradient norm, policy entropy, validation accuracy, and
        Pass@4 over training. \method establishes an early lead over PPO
        in both performance metrics and retains it through the final checkpoint.
    }
    \vspace{-1.5em}
    \label{fig:training_dynamics}
\end{figure}

\begin{table}[t]
    \centering
    \caption{
        \textbf{In-domain and out-of-domain results with $n=1$.}
        All entries are accuracies (\%).
        Orange parenthetical values show the absolute improvement over PPO
        in percentage points.
        Bold and underlined values indicate the best and second-best
        results within each model group, respectively.
    }
    \label{tab:main_results_n1}
    \small
    \setlength{\tabcolsep}{2.45pt}
    \renewcommand{\arraystretch}{1.25}
    \resizebox{\linewidth}{!}{%
    \begin{tabular}{@{}lcccccc|cccc@{}}
        \toprule
        \rowcolor{tableheader}
        \textbf{Method}
        & \multicolumn{6}{c|}{\textbf{In-Domain Performance}}
        & \multicolumn{4}{c}{\textbf{Out-of-Domain Performance}} \\
        \rowcolor{tableheader}
        & \textbf{AIME24/25/26}
        & \textbf{AMC}
        & \textbf{MATH500}
        & \textbf{Minerva}
        & \textbf{Olympiad}
        & \textbf{Avg.}
        & \textbf{ARC-C}
        & \textbf{AGI-Eval}
        & \textbf{BBH}
        & \textbf{OOD Avg.} \\
        \midrule
        \rowcolor{tablegroup}
        \multicolumn{11}{c}{\textbf{Qwen3-8B-Base}} \\
        \midrule
        Base
        & 8.2/6.8/6.4 & 33.8 & 54.3
        & 24.9 & 27.2 & 23.1
        & 62.4 & 44.2 & 27.8 & 44.8 \\
        PPO
        & 11.7/\underline{15.0}/\underline{15.0} & \underline{56.9} & 77.9
        & \underline{45.9} & 39.2 & \underline{37.4}
        & 91.4 & \underline{67.1} & 42.0 & \underline{66.8} \\
        SPPO
        & \underline{15.8}/10.8/\underline{15.0} & 56.3 & \underline{78.2}
        & 43.4 & \underline{41.3} & 37.3
        & \underline{91.8} & 66.0 & 40.4 & 66.1 \\
        \rowcolor{methodlight}
        \textcolor{methodtext}{\textbf{\method}}
        & \textbf{23.3}/\textbf{20.0}/\textbf{20.8}
        & \textbf{68.1} & \textbf{84.9}
        & \textbf{46.3} & \textbf{50.9}
        & \textbf{44.9}\,\ppoGain{+7.5}
        & \textbf{93.0} & \textbf{71.8}
        & \textbf{46.8} & \textbf{70.5} \\
        % \rowcolor{methodstrong}
        % \textcolor{methodtext}{\textbf{\method + PI}}
        % & \underline{22.5}/\underline{19.2}/\underline{16.7}
        % & \textbf{68.8} & \underline{85.6}
        % & 44.9 & \underline{47.2}
        % & \underline{43.5}\,\ppoGain{+8.0}
        % & \underline{62.8} & \textbf{72.2}
        % & \underline{43.7} & \underline{59.6} \\
        \midrule
        \rowcolor{tablegroup}
        \multicolumn{11}{c}{\textbf{Qwen3-4B}} \\
        \midrule
        Base
        & 19.2/24.2/17.5 & 57.5 & 79.4
        & 47.2 & 43.5 & 41.2
        & 91.3 & 74.0 & 46.6 & 70.6 \\
        PPO
        & \underline{44.2}/\underline{35.0}/34.2
        & 79.4 & 85.5
        & 47.2 & 52.7
        & 54.0
        & \underline{91.3} & 78.7
        & 50.6 & 73.5 \\
        SPPO
        & 40.8/32.5/\underline{35.0} & \underline{85.0} & \underline{87.1}
        & \underline{50.5} & \underline{56.9} & \underline{55.4}
        & \textbf{91.6} & \underline{79.4} & \textbf{52.6} & \textbf{74.5} \\
        \rowcolor{methodlight}
        \textcolor{methodtext}{\textbf{\method}}
        & \textbf{60.8}/\textbf{47.5}/\textbf{44.2}
        & \textbf{90.6} & \textbf{92.3}
        & \textbf{51.0} & \textbf{62.4}
        & \textbf{64.1}\,\ppoGain{+10.1}
        & 90.3 & \textbf{79.9}
        & \underline{51.1} & \underline{73.8} \\
        \bottomrule
    \end{tabular}%
    }
\end{table}

% Keep the results heading, wrapped plot, and caption on the same page.
\Needspace{20\baselineskip}
\subsection{Results}
\label{sec:main_results}

\begin{wrapfigure}{r}{0.4\linewidth}
    \centering
    \captionsetup{skip=4pt}
    \vspace{-1.5em}
    \includegraphics[width=\linewidth]{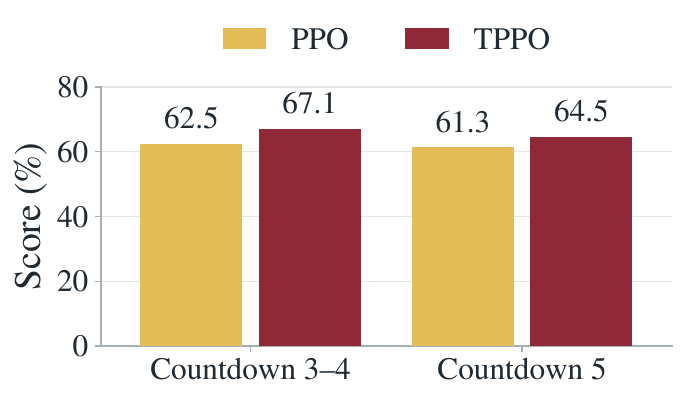}
    \caption{\small
        \textbf{Countdown task on Qwen2.5-Math-1.5B.}
        Trained on problems with three or four input numbers.
    }
    \vspace{-0.5em}
    \label{fig:countdown_results}
\end{wrapfigure}

\paragraph{RQ1: \method improves policy performance across model sizes and tasks.}
Tables~\ref{tab:main_results} and~\ref{tab:main_results_n1} show consistent
in-domain gains over PPO across model sizes and rollout budgets.
With $n=8$, \method improves average accuracy by 7.0 and 8.4 percentage
points on Qwen3-4B-Base and Qwen3-8B-Base, respectively, outperforming
baselines.
The gains persist with $n=1$, including an average improvement of +10.1 points on Qwen3-4B.
Figure~\ref{fig:training_dynamics} shows an early and sustained lead in validation accuracy and Pass@4, with smaller actor gradient norms and
slower entropy decay suggesting more stable updates and less premature
policy collapse.
Results for Llama3.2-3B-Instruct on mathematical reasoning are reported in
Appendix~\ref{app:additional_results}, Figure~\ref{fig:llama32_results}.
The gains also extend to Countdown (+4.6 on 3--4 input numbers and +3.2 on 5 input numbers) with Qwen2.5-Math-1.5B (Figure~\ref{fig:countdown_results}), and to scientific coding (+7.0 on average and +15.5 on LiveCodeBench Code Gen.v6) with Qwen3.5-9B
(Table~\ref{tab:science_code_results}).

\begin{wraptable}{R}{0.50\textwidth}
    \centering
    \captionsetup{skip=4pt}
    \caption{\small
        \textbf{Coding results on Qwen3.5-9B.}
        Accuracy (\%) averaged over three evaluation runs. Orange parenthetical
        values show the improvement over PPO in percentage points.
        HumanEvalFix averages Python, Rust, and JavaScript.
    }
    \label{tab:science_code_results}
    \small
    \setlength{\tabcolsep}{3pt}
    \renewcommand{\arraystretch}{1.1}
    \begin{tabular*}{\linewidth}{@{\extracolsep{\fill}}lcc>{\columncolor{methodlight}}c@{}}
        \toprule
        \rowcolor{tableheader}
        \textbf{Benchmark} & \textbf{Base} & \textbf{PPO}
        & \textcolor{methodtext}{\textbf{\method}} \\
        \midrule
        SciCode & 29.3 & 32.6 & \textbf{35.0}\,\ppoGain{+2.4} \\
        HumanEval & 94.7 & 93.5 & \textbf{96.5}\,\ppoGain{+3.0} \\
        HumanEvalFix & 58.3 & 70.1 & \textbf{75.5}\,\ppoGain{+5.4} \\
        LCB Code Gen.v6 & 61.9 & 57.6 & \textbf{73.1}\,\ppoGain{+15.5} \\
        APPS Introductory & 69.1 & 72.3 & \textbf{80.8}\,\ppoGain{+8.5} \\
        \bottomrule
    \end{tabular*}
    \vspace{-0.5em}
\end{wraptable}

\begin{figure}[t]
    \centering
    \captionsetup{skip=4pt}
    \begin{minipage}[t]{0.49\linewidth}
        \vspace{0pt}
        \centering
        \includegraphics[width=\linewidth]{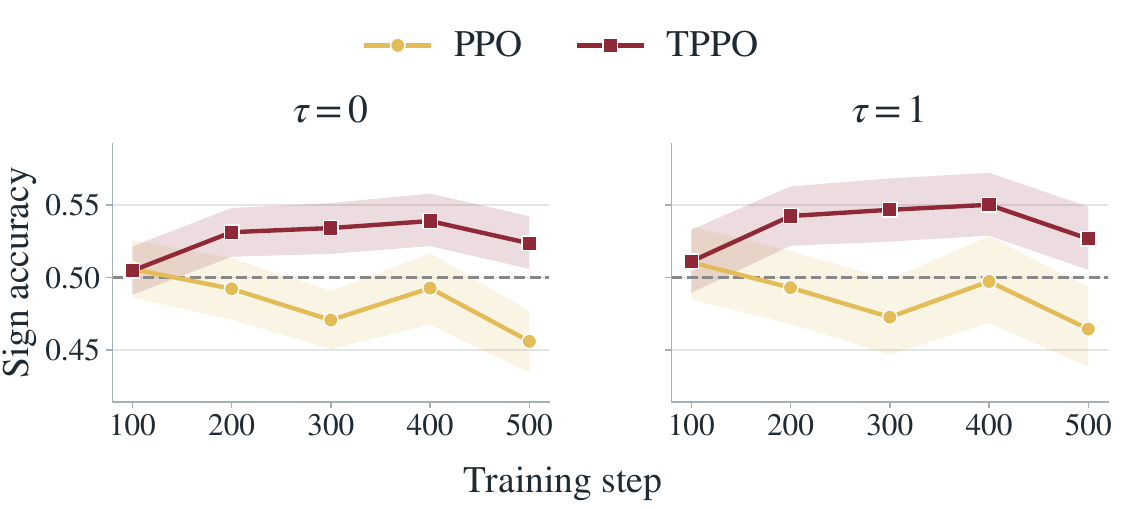}
        \caption{\small
            \textbf{Value evolution tracking.} Sign agreement between critic
            predictions and MC-estimated value changes for $\mathrm{SNR}>\tau$.
            Results use Qwen3-8B-Base mathematical reasoning with $n=8$.
        }
        \label{fig:value_changes_metrics}
    \end{minipage}\hfill
    \begin{minipage}[t]{0.49\linewidth}
        \vspace{0pt}
        \centering
        \includegraphics[width=\linewidth]{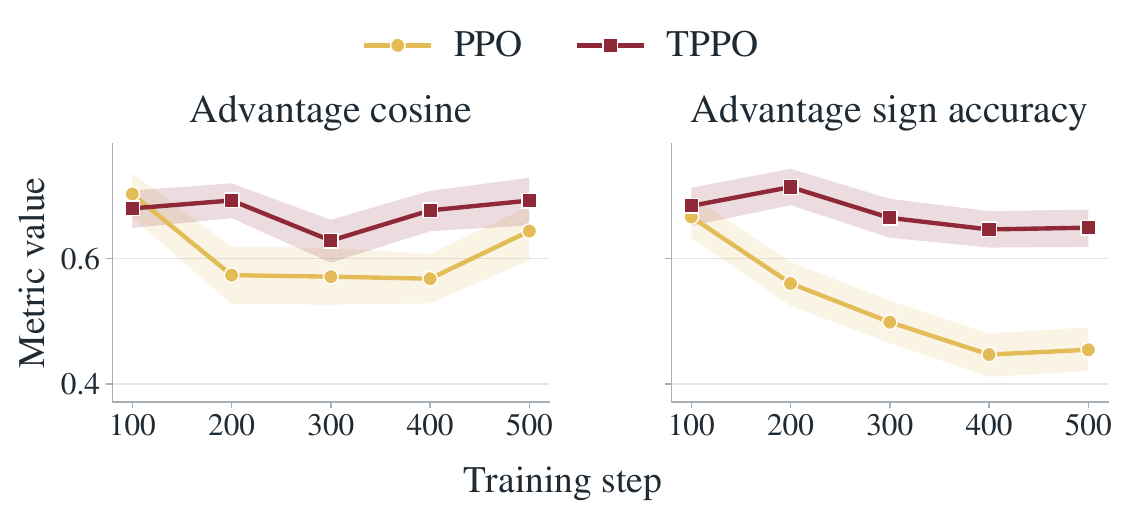}
        \caption{\small
            \textbf{Credit assignment.} Cosine similarity and sign agreement
            with GAE advantages computed from MC values. Results use
            Qwen3-8B-Base mathematical reasoning with $n=8$.
        }
        \label{fig:advantage_metrics_training}
    \end{minipage}
    \vspace{-1.5em}
\end{figure}

\paragraph{RQ2: \method better tracks value evolution during training.}
Figure~\ref{fig:value_changes_metrics} compares critic-predicted and MC-estimated
local value evolution between adjacent anchors. We evaluate under two SNR
thresholds ($\tau=0,1$) to check whether the comparison is robust to
filtering out noisier estimates of local value evolution (Appendix~\ref{app:diagnostic_metrics}).
Starting from similar accuracy, \method maintains stronger directional
agreement as training progresses, while PPO's agreement declines.
Across the last four checkpoints, the gains are 3.9--6.8 percentage
points for $\tau=0$ and 4.9--7.4 for $\tau=1$, with the corresponding
95\% bootstrap intervals for the gains above zero.
These results show improved tracking of the direction of local value evolution under
both SNR thresholds.
Appendix Figure~\ref{fig:rollout_source_sign_accuracy} additionally compares
both critics on the same actor's rollouts to control for rollout source.
\looseness=-1
\vspace{-.5em}

\paragraph{RQ3: \method improves the quality of credit assignment.}
Figure~\ref{fig:advantage_metrics_training} evaluates GAE advantages ($\lambda=1$) against a reference computed by substituting MC estimates for critic values (details are in Appendix~\ref{app:diagnostic_metrics}).
After the initial checkpoint, \method consistently achieves higher cosine similarity and sign agreement with this reference.
Across the last four checkpoints, advantage sign accuracy improves by 15.3 - 19.9 percentage points, with 95\% bootstrap intervals for these
gains above zero. Together with RQ2, these results show that improved
tracking value evolution accompanies advantage estimates that agree more closely with the MC reference.

\section{Discussion}
\label{sec:discussion}

\begin{figure}[t]
    \centering
    \includegraphics[width=1.0\linewidth]
    {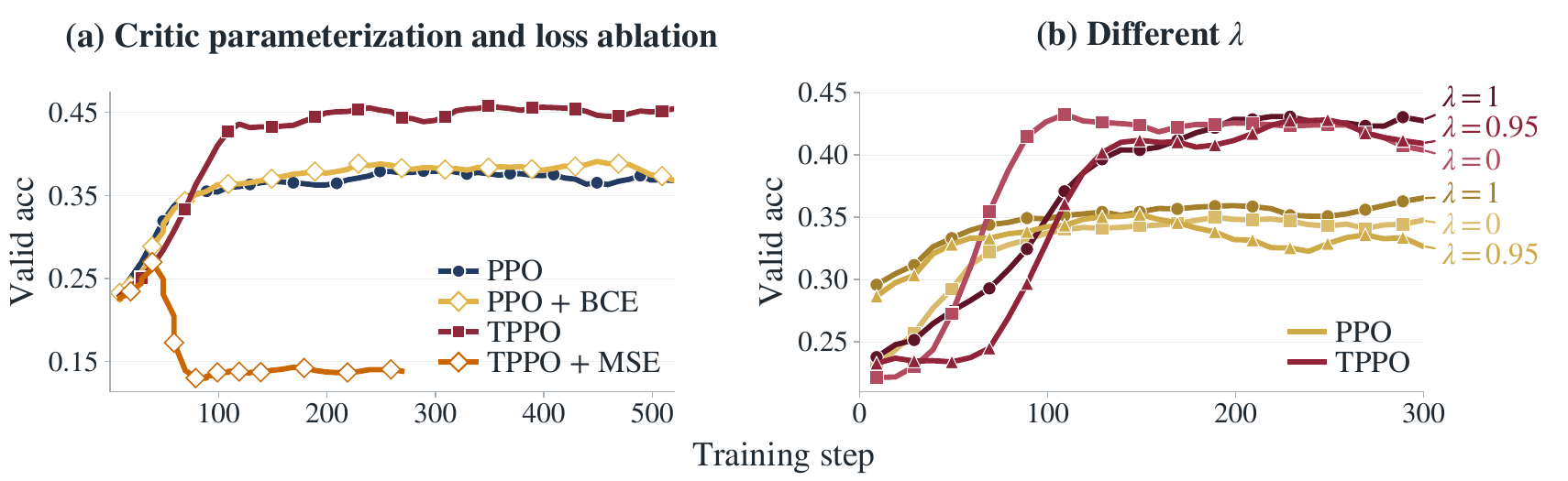}
    \caption{\small
        \textbf{Critic ablations and actor GAE coefficients.}
        Validation accuracy for (a) critic parameterization and loss variants,
        and (b) PPO and \method with $\lambda_{\mathrm{actor}}\in\{1,0.95,0\}$.
        Settings for (b) are given in Appendix~\ref{app:actor_gae_settings}.
    }
    \label{fig:ablate_lambda}
    \vspace{-1.2em}
\end{figure}

\paragraph{Ablations}
To test whether the gains come from the loss alone, we train a standard
PPO critic with BCE (PPO + BCE), applying a sigmoid to each position's
scalar output without accumulating outputs across positions. We also replace
BCE with MSE in \method (\method + MSE), keeping its cumulative
parameterization.
\label{sec:controlled_ablations}
As shown in Figure~\ref{fig:ablate_lambda} (a), PPO + BCE improves over PPO, but \method reaches a higher validation accuracy and maintains the
gap.
The \method + MSE variant deteriorates early and remains well below the
other variants. The gains therefore mainly come from modeling value evolution
with the cumulative parameterization, with BCE providing an additional benefit.

\vspace{-0.7em}
\begin{wrapfigure}{r}{0.5\linewidth}
    \centering
    \captionsetup{skip=4pt}
    \vspace{-1em}
    \includegraphics[width=\linewidth]{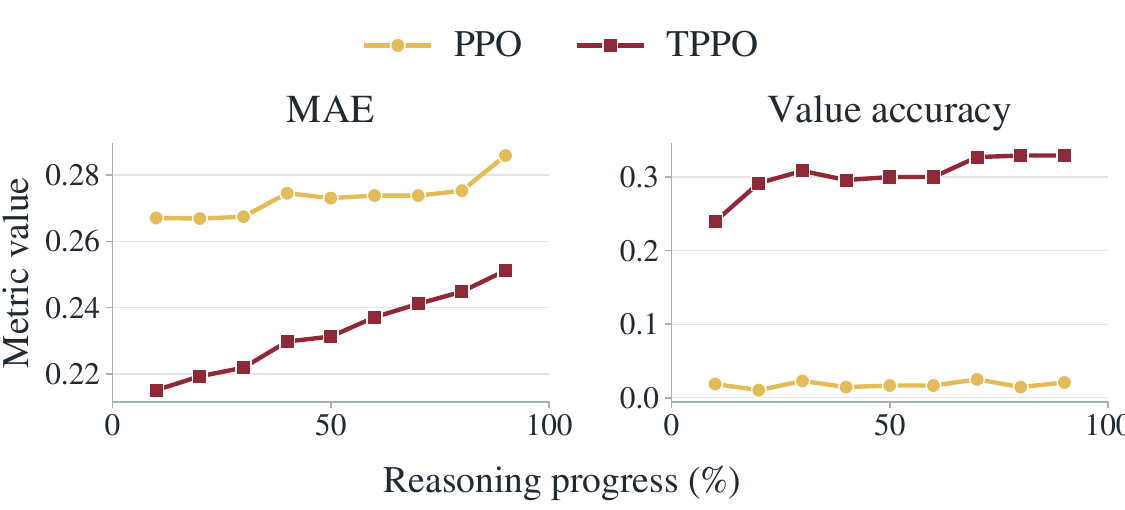}
    \caption{\small \textbf{Value estimates on test data.} Error and accuracy against MC estimates at different reasoning positions.}
    \label{fig:value_estimates_test}
    \vspace{-1em}
\end{wrapfigure}
\paragraph{Modeling value evolution improves value estimates.}
Following the usual interpretation of a critic as an estimate of the expected
terminal return from a prefix~\citep{kazemnejad2024vineppo}, we compare the
reconstructed values with Monte Carlo (MC) estimates. Figure~\ref{fig:value_estimates_test}
shows this comparison on AIME responses. \method has lower mean absolute
error and higher value accuracy at every evaluated position. Its error grows
towards the end of a response but remains below PPO's, while its accuracy stays
higher throughout. This supports the view
that learning value evolution improves the critic's value estimates, which in
turn provide the baseline and local changes used by GAE.
\looseness=-1

\vspace{-0.7em}
\paragraph{Modeling value evolution improves credit assignment for different GAE $\lambda$.}
We vary the actor's GAE coefficient $\lambda$, which controls how far
advantage estimates propagate information from later TD residuals, to test
whether learning value evolution remains useful beyond the setting used in our
main experiments.
In this experiment, we vary only the $\lambda$ used to compute the actor's
GAE advantages, keeping the critic's $\lambda$ fixed at 1.
Under terminal rewards and $\gamma=1$, $\lambda=0$ uses the local value change
at nonterminal transitions, while $0<\lambda<1$ geometrically weights subsequent
value changes and the terminal TD residual. At $\lambda=1$, these terms
telescope to the terminal reward minus the current value
(Appendix~\ref{app:value_change_credit}). Better modeling of value evolution can
therefore improve credit assignment across these regimes. Our $\lambda=0$
experiment applies this construction at the segment level.
Figure~\ref{fig:ablate_lambda}(b)
shows that \method retains its performance advantage over PPO in every
evaluated setting; the exact training settings are given in
Appendix~\ref{app:actor_gae_settings}.

% \begin{figure}[t]
%     \centering
% \begin{minipage}[t]{0.54\linewidth}
%     \vspace{0pt}
%     \centering
%     \captionsetup{skip=4pt}
%     \includegraphics[width=\linewidth]{sources/figures/outputs/04/test/04_value_quality_test_absolute.pdf}
%     \caption{\small \textbf{Value estimates on test data.} Error and accuracy against MC estimates at different reasoning positions.}
%     \label{fig:value_estimates_test}
% \end{minipage}\hfill
% \begin{minipage}[t]{0.44\linewidth}
%     \vspace{0pt}
%     \centering
%     \captionsetup{skip=4pt}
%     \includegraphics[width=\linewidth]{sources/figures/8b_valid_acc_lambda_comparison.pdf}
%     \caption{\small \textbf{PPO and \method under different values of $\lambda$.}}
%     \label{fig:actor_gae_lambda}
% \end{minipage}
% \vspace{-2.0em}
% \end{figure}

\section{Conclusion}
\label{sec:conclusion}

We presented \method, a cumulative parameterization for modeling value evolution in language-model reinforcement learning. Our analysis of successive prefix values reveals a Bayesian decomposition with additive updates in log-odds space.
\method uses this structure to accumulate position-wise scalar outputs and learns these outputs through terminal-outcome supervision on the reconstructed values, which are then used by PPO.
Experiments on mathematical reasoning tasks show consistent improvements over PPO across two Qwen3 model sizes, as well as gains on additional tasks and a different model family. The critic analyses show stronger agreement with Monte Carlo values and better alignment with MC-based reference advantages, while the ablations show that the cumulative parameterization and BCE training are both important.
These results suggest that modeling how value evolves along a
reasoning trace can improve credit assignment without additional rollout sampling or process-level labels.

\section*{Acknowledgments}

This work was supported by the Shanghai Artificial Intelligence Laboratory. We are grateful to the authors and open-source communities whose work made this project possible.
\bibliography{sources/reference}

@article{lewkowycz2022solving,
  title={Solving quantitative reasoning problems with language models},
  author={Lewkowycz, Aitor and Andreassen, Anders and Dohan, David and Dyer, Ethan and Michalewski, Henryk and Ramasesh, Vinay and Slone, Ambrose and Anil, Cem and Schlag, Imanol and Gutman-Solo, Theo and others},
  journal={Advances in neural information processing systems},
  volume={35},
  pages={3843--3857},
  year={2022}
}

@misc{slime_github,
  author       = {Zilin Zhu and Chengxing Xie and Xin Lv and slime Contributors},
  title        = {slime: An LLM post-training framework for RL Scaling},
  year         = {2025},
  howpublished = {\url{https://github.com/THUDM/slime}},
  note         = {GitHub repository. Corresponding author: Xin Lv},
  urldate      = {2025-06-19}
}

@article{yang2025qwen3,
  title={Qwen3 technical report},
  author={Yang, An and Li, Anfeng and Yang, Baosong and Zhang, Beichen and Hui, Binyuan and Zheng, Bo and Yu, Bowen and Gao, Chang and Huang, Chengen and Lv, Chenxu and others},
  journal={arXiv preprint arXiv:2505.09388},
  year={2025}
}

@article{hendrycks2021measuring,
  title={Measuring mathematical problem solving with the math dataset},
  author={Hendrycks, Dan and Burns, Collin and Kadavath, Saurav and Arora, Akul and Basart, Steven and Tang, Eric and Song, Dawn and Steinhardt, Jacob},
  journal={arXiv preprint arXiv:2103.03874},
  year={2021}
}

@article{li2024numinamath,
  title={Numinamath: The largest public dataset in ai4maths with 860k pairs of competition math problems and solutions},
  author={Li, Jia and Beeching, Edward and Tunstall, Lewis and Lipkin, Ben and Soletskyi, Roman and Huang, Shengyi and Rasul, Kashif and Yu, Longhui and Jiang, Albert Q and Shen, Ziju and others},
  journal={Hugging Face repository},
  volume={13},
  number={9},
  pages={9},
  year={2024}
}

@inproceedings{he2024olympiadbench,
  title={Olympiadbench: A challenging benchmark for promoting agi with olympiad-level bilingual multimodal scientific problems},
  author={He, Chaoqun and Luo, Renjie and Bai, Yuzhuo and Hu, Shengding and Thai, Zhen and Shen, Junhao and Hu, Jinyi and Han, Xu and Huang, Yujie and Zhang, Yuxiang and others},
  booktitle={Proceedings of the 62nd Annual Meeting of the Association for Computational Linguistics (Volume 1: Long Papers)},
  pages={3828--3850},
  year={2024}
}

@misc{Polaris2025,
    title = {POLARIS: A Post-Training Recipe for Scaling Reinforcement Learning on Advanced Reasoning Models},
    url = {https://hkunlp.github.io/blog/2025/Polaris},
    author = {An, Chenxin and Xie, Zhihui and Li, Xiaonan and Li, Lei and Zhang, Jun and Gong, Shansan and Zhong, Ming and Xu, Jingjing and Qiu, Xipeng and Wang, Mingxuan and Kong, Lingpeng},
    year = {2025}
}

@article{schulman2017proximal,
  title={Proximal policy optimization algorithms},
  author={Schulman, John and Wolski, Filip and Dhariwal, Prafulla and Radford, Alec and Klimov, Oleg},
  journal={arXiv preprint arXiv:1707.06347},
  year={2017}
}

@article{schulman2015high,
  title={High-dimensional continuous control using generalized advantage estimation},
  author={Schulman, John and Moritz, Philipp and Levine, Sergey and Jordan, Michael and Abbeel, Pieter},
  journal={arXiv preprint arXiv:1506.02438},
  year={2015}
}

@book{sutton1998reinforcement,
  title={Reinforcement learning: An introduction},
  author={Sutton, Richard S and Barto, Andrew G and others},
  volume={1},
  number={1},
  year={1998},
  publisher={MIT press Cambridge}
}

@article{shao2024deepseekmath,
  title={Deepseekmath: Pushing the limits of mathematical reasoning in open language models},
  author={Shao, Zhihong and Wang, Peiyi and Zhu, Qihao and Xu, Runxin and Song, Junxiao and Bi, Xiao and Zhang, Haowei and Zhang, Mingchuan and Li, YK and Wu, Yang and others},
  journal={arXiv preprint arXiv:2402.03300},
  year={2024}
}

@article{yu2026dapo,
  title={Dapo: An open-source llm reinforcement learning system at scale},
  author={Yu, Qiying and Zhang, Zheng and Zhu, Ruofei and Yuan, Yufeng and Zuo, Xiaochen and Yue, Yu and Dai, Weinan and Fan, Tiantian and Liu, Gaohong and Liu, Lingjun and others},
  journal={Advances in Neural Information Processing Systems},
  volume={38},
  pages={113222--113244},
  year={2026}
}

@article{kazemnejad2024vineppo,
  title={Vineppo: Refining credit assignment in rl training of llms},
  author={Kazemnejad, Amirhossein and Aghajohari, Milad and Portelance, Eva and Sordoni, Alessandro and Reddy, Siva and Courville, Aaron and Roux, Nicolas Le},
  journal={arXiv preprint arXiv:2410.01679},
  year={2024}
}

@article{farebrother2024stop,
  title={Stop regressing: Training value functions via classification for scalable deep rl},
  author={Farebrother, Jesse and Orbay, Jordi and Vuong, Quan and Ta{\"\i}ga, Adrien Ali and Chebotar, Yevgen and Xiao, Ted and Irpan, Alex and Levine, Sergey and Castro, Pablo Samuel and Faust, Aleksandra and others},
  journal={arXiv preprint arXiv:2403.03950},
  year={2024}
}

@article{zhou2026start,
  title={Start Classifying: Categorical Critics for LLM Reinforcement Learning},
  author={Zhou, Zhijian and Li, Long and Zhang, Xuan and Liu, Zongkai and Qin, Yulei and Li, Ke and Sun, Xing and Tan, Xiaoyu and Qu, Chao and Qi, Yuan},
  journal={arXiv preprint arXiv:2608.02181},
  year={2026}
}

@article{cui2025process,
  title={Process reinforcement through implicit rewards},
  author={Cui, Ganqu and Yuan, Lifan and Wang, Zefan and Wang, Hanbin and Zhang, Yuchen and Chen, Jiacheng and Li, Wendi and He, Bingxiang and Fan, Yuchen and Yu, Tianyu and others},
  journal={arXiv preprint arXiv:2502.01456},
  year={2025}
}

@inproceedings{bellemare2017distributional,
  title={A distributional perspective on reinforcement learning},
  author={Bellemare, Marc G and Dabney, Will and Munos, R{\'e}mi},
  booktitle={International conference on machine learning},
  pages={449--458},
  year={2017},
  organization={Pmlr}
}

@article{hou2506treerl,
  title={Treerl: Llm reinforcement learning with on-policy tree search, 2025},
  author={Hou, Zhenyu and Hu, Ziniu and Li, Yujiang and Lu, Rui and Tang, Jie and Dong, Yuxiao},
  journal={URL https://arxiv. org/abs/2506.11902}
}

@article{qu2025optimizing,
  title={Optimizing test-time compute via meta reinforcement fine-tuning},
  author={Qu, Yuxiao and Yang, Matthew YR and Setlur, Amrith and Tunstall, Lewis and Beeching, Edward Emanuel and Salakhutdinov, Ruslan and Kumar, Aviral},
  journal={arXiv preprint arXiv:2503.07572},
  year={2025}
}

@inproceedings{lightman2024let,
  title={Let's verify step by step},
  author={Lightman, Hunter and Kosaraju, Vineet and Burda, Yuri and Edwards, Harrison and Baker, Bowen and Lee, Teddy and Leike, Jan and Schulman, John and Sutskever, Ilya and Cobbe, Karl},
  booktitle={International Conference on Learning Representations},
  volume={2024},
  pages={39578--39601},
  year={2024}
}

@inproceedings{wang2024math,
  title={Math-shepherd: Verify and reinforce llms step-by-step without human annotations},
  author={Wang, Peiyi and Li, Lei and Shao, Zhihong and Xu, Runxin and Dai, Damai and Li, Yifei and Chen, Deli and Wu, Yu and Sui, Zhifang},
  booktitle={Proceedings of the 62nd Annual Meeting of the Association for Computational Linguistics (Volume 1: Long Papers)},
  pages={9426--9439},
  year={2024}
}

@inproceedings{setlur2025rewarding,
  title={Rewarding progress: Scaling automated process verifiers for llm reasoning},
  author={Setlur, Amrith and Nagpal, Chirag and Fisch, Adam and Geng, Xinyang and Eisenstein, Jacob and Agarwal, Rishabh and Agarwal, Alekh and Berant, Jonathan and Kumar, Aviral},
  booktitle={International Conference on Learning Representations},
  volume={2025},
  pages={60808--60838},
  year={2025}
}

@article{li2025treepo,
  title={Treepo: Bridging the gap of policy optimization and efficacy and inference efficiency with heuristic tree-based modeling},
  author={Li, Yizhi and Gu, Qingshui and Wen, Zhoufutu and Li, Ziniu and Xing, Tianshun and Guo, Shuyue and Zheng, Tianyu and Zhou, Xin and Qu, Xingwei and Zhou, Wangchunshu and others},
  journal={arXiv preprint arXiv:2508.17445},
  year={2025}
}

@article{yuan2025s,
  title={What's Behind PPO's Collapse in Long-CoT? Value Optimization Holds the Secret},
  author={Yuan, Yufeng and Yue, Yu and Zhu, Ruofei and Fan, Tiantian and Yan, Lin},
  journal={arXiv preprint arXiv:2503.01491},
  year={2025}
}

@inproceedings{wang2026sppo,
  title={Sppo: Sequence-level ppo for long-horizon reasoning tasks},
  author={Wang, Tianyi and Li, Yixia and Li, Long and Chen, Yibiao and Huang, Shaohan and Chen, Yun and Li, Peng and Liu, Yang and Chen, Guanhua},
  booktitle={Proceedings of the 64th Annual Meeting of the Association for Computational Linguistics (Volume 1: Long Papers)},
  pages={7831--7853},
  year={2026}
}

@article{guo2025deepseek,
  title={Deepseek-r1: Incentivizing reasoning capability in llms via reinforcement learning},
  author={Guo, Daya and Yang, Dejian and Zhang, Haowei and Song, Junxiao and Wang, Peiyi and Zhu, Qihao and Xu, Runxin and Zhang, Ruoyu and Ma, Shirong and Bi, Xiao and others},
  journal={arXiv preprint arXiv:2501.12948},
  year={2025}
}

@article{jaech2024openai,
  title={Openai o1 system card},
  author={Jaech, Aaron and Kalai, Adam and Lerer, Adam and Richardson, Adam and El-Kishky, Ahmed and Low, Aiden and Helyar, Alec and Madry, Aleksander and Beutel, Alex and Carney, Alex and others},
  journal={arXiv preprint arXiv:2412.16720},
  year={2024}
}

@inproceedings{zhang2025lessons,
  title={The lessons of developing process reward models in mathematical reasoning},
  author={Zhang, Zhenru and Zheng, Chujie and Wu, Yangzhen and Zhang, Beichen and Lin, Runji and Yu, Bowen and Liu, Dayiheng and Zhou, Jingren and Lin, Junyang},
  booktitle={Findings of the Association for Computational Linguistics: ACL 2025},
  pages={10495--10516},
  year={2025}
}

@inproceedings{
chen2026hista,
title={Hista and Numca: Estimate State Value Effectively for Large Language Model Reinforcement Learning},
author={Zizhe Chen and Jiqian Dong and Yizhou Tian and Garry YANG and Yongqiang Chen and Zhitang Chen and James Cheng},
booktitle={Forty-third International Conference on Machine Learning},
year={2026},
url={https://openreview.net/forum?id=ff4b0ELdU0}
}

@inproceedings{zhong2024agieval,
  title={Agieval: A human-centric benchmark for evaluating foundation models},
  author={Zhong, Wanjun and Cui, Ruixiang and Guo, Yiduo and Liang, Yaobo and Lu, Shuai and Wang, Yanlin and Saied, Amin and Chen, Weizhu and Duan, Nan},
  booktitle={Findings of the association for computational linguistics: NAACL 2024},
  pages={2299--2314},
  year={2024}
}

@inproceedings{suzgun2023challenging,
  title={Challenging big-bench tasks and whether chain-of-thought can solve them},
  author={Suzgun, Mirac and Scales, Nathan and Sch{\"a}rli, Nathanael and Gehrmann, Sebastian and Tay, Yi and Chung, Hyung Won and Chowdhery, Aakanksha and Le, Quoc and Chi, Ed H and Zhou, Denny and others},
  booktitle={Findings of the Association for Computational Linguistics: ACL 2023},
  pages={13003--13051},
  year={2023}
}

@article{yue2025vapo,
  title={Vapo: Efficient and reliable reinforcement learning for advanced reasoning tasks},
  author={Yue, Yu and Yuan, Yufeng and Yu, Qiying and Zuo, Xiaochen and Zhu, Ruofei and Xu, Wenyuan and Chen, Jiaze and Wang, Chengyi and Fan, TianTian and Du, Zhengyin and others},
  journal={arXiv preprint arXiv:2504.05118},
  year={2025}
}

@inproceedings{trung2024reft,
  title={Reft: Reasoning with reinforced fine-tuning},
  author={Trung, Luong and Zhang, Xinbo and Jie, Zhanming and Sun, Peng and Jin, Xiaoran and Li, Hang},
  booktitle={Proceedings of the 62nd Annual Meeting of the Association for Computational Linguistics (Volume 1: Long Papers)},
  pages={7601--7614},
  year={2024}
}

@article{ouyang2022training,
  title={Training language models to follow instructions with human feedback},
  author={Ouyang, Long and Wu, Jeffrey and Jiang, Xu and Almeida, Diogo and Wainwright, Carroll and Mishkin, Pamela and Zhang, Chong and Agarwal, Sandhini and Slama, Katarina and Ray, Alex and others},
  journal={Advances in neural information processing systems},
  volume={35},
  pages={27730--27744},
  year={2022}
}

@misc{li2026sciwalkersynthesizingscientificcoding,
      title={SciWalker: Synthesizing Scientific Coding Problems with Operator Graphs and Execution Feedback}, 
      author={Chenxi Li and Wenxuan Zeng and Yun Luo and Fangchen Yu and Peng Ye and Yu Cheng and Jun Zhang},
      year={2026},
      eprint={2609.30054},
      archivePrefix={arXiv},
      primaryClass={cs.AI},
      url={https://arxiv.org/abs/2609.30054}, 
}

@inproceedings{SciCode,
 author = {Tian, Minyang and Gao, Luyu and Zhang, Shizhuo Dylan and Chen, Xinan and Fan, Cunwei and Guo, Xuefei and Haas, Roland and Ji, Pan and Krongchon, Kittithat and Li, Yao and Liu, Shengyan and Luo, Di and Ma, Yutao and Tong, Hao and Trinh, Kha and Tian, Chenyu and Wang, Zihan and Wu, Bohao and Xiong, Yanyu and Yin, Shengzhu and Zhu, Minhui and Lieret, Kilian and Lu, Yanxin and Liu, Genglin and Du, Yufeng and Tao, Tianhua and Press, Ofir and Callan, Jamie and Huerta, Eliu and Peng, Hao},
 booktitle = {Advances in Neural Information Processing Systems},
 doi = {10.52202/079017-0963},
 editor = {A. Globerson and L. Mackey and D. Belgrave and A. Fan and U. Paquet and J. Tomczak and C. Zhang},
 pages = {30624--30650},
 publisher = {Curran Associates, Inc.},
 title = {SciCode: A Research Coding Benchmark Curated by Scientists},
 url = {https://proceedings.neurips.cc/paper_files/paper/2024/file/36850592258c8c41cecdaa3dea5ff7de-Paper-Datasets_and_Benchmarks_Track.pdf},
 volume = {37},
 year = {2024}
}

@misc{humaneval,
      title={Evaluating Large Language Models Trained on Code}, 
      author={Mark Chen and Jerry Tworek and Heewoo Jun and Qiming Yuan and Henrique Ponde de Oliveira Pinto and Jared Kaplan and Harri Edwards and Yuri Burda and Nicholas Joseph and Greg Brockman and Alex Ray and Raul Puri and Gretchen Krueger and Michael Petrov and Heidy Khlaaf and Girish Sastry and Pamela Mishkin and Brooke Chan and Scott Gray and Nick Ryder and Mikhail Pavlov and Alethea Power and Lukasz Kaiser and Mohammad Bavarian and Clemens Winter and Philippe Tillet and Felipe Petroski Such and Dave Cummings and Matthias Plappert and Fotios Chantzis and Elizabeth Barnes and Ariel Herbert-Voss and William Hebgen Guss and Alex Nichol and Alex Paino and Nikolas Tezak and Jie Tang and Igor Babuschkin and Suchir Balaji and Shantanu Jain and William Saunders and Christopher Hesse and Andrew N. Carr and Jan Leike and Josh Achiam and Vedant Misra and Evan Morikawa and Alec Radford and Matthew Knight and Miles Brundage and Mira Murati and Katie Mayer and Peter Welinder and Bob McGrew and Dario Amodei and Sam McCandlish and Ilya Sutskever and Wojciech Zaremba},
      year={2021},
      eprint={2107.03374},
      archivePrefix={arXiv},
      primaryClass={cs.LG},
      url={https://arxiv.org/abs/2107.03374}, 
}

@inproceedings{HumanEvalFix,
 author = {Muennighoff, Niklas and Liu, Qian and Zebaze, Armel and Zheng, Qinkai and Hui, Binyuan and Zhuo, Terry Yue and Singh, Swayam and Tang, Xiangru and Von Werra, Leandro and Longpre, Shayne},
 booktitle = {International Conference on Learning Representations},
 editor = {B. Kim and Y. Yue and S. Chaudhuri and K. Fragkiadaki and M. Khan and Y. Sun},
 pages = {7604--7623},
 title = {OctoPack: Instruction Tuning Code Large Language Models},
 url = {https://proceedings.iclr.cc/paper_files/paper/2024/file/1ec299a5229034141e58aeded0d0b9de-Paper-Conference.pdf},
 volume = {2024},
 year = {2024}
}

@inproceedings{LiveCodeBench,
 author = {Jain, Naman and Han and Gu, Alex and Li, Wen-Ding and Yan, Fanjia and Zhang, Tianjun and Wang, Sida and Solar-Lezama, Armando and Sen, Koushik and Stoica, Ion},
 booktitle = {International Conference on Learning Representations},
 editor = {Y. Yue and A. Garg and N. Peng and F. Sha and R. Yu},
 pages = {58791--58831},
 title = {LiveCodeBench: Holistic and Contamination Free Evaluation of Large Language Models for Code},
 url = {https://proceedings.iclr.cc/paper_files/paper/2025/file/94074dd5a072d28ff75a76dabed43767-Paper-Conference.pdf},
 volume = {2025},
 year = {2025}
}

@inproceedings{APPS,
 author = {Hendrycks, Dan and Basart, Steven and Kadavath, Saurav and Mazeika, Mantas and Arora, Akul and Guo, Ethan and Burns, Collin and Puranik, Samir and He, Horace and Song, Dawn and Steinhardt, Jacob},
 booktitle = {Proceedings of the Neural Information Processing Systems Track on Datasets and Benchmarks},
 editor = {J. Vanschoren and S. Yeung},
 pages = {},
 title = {Measuring Coding Challenge Competence With APPS},
 url = {https://datasets-benchmarks-proceedings.neurips.cc/paper_files/paper/2021/file/c24cd76e1ce41366a4bbe8a49b02a028-Paper-round2.pdf},
 volume = {1},
 year = {2021}
}

@article{clark2018think,
  title={Think you have solved question answering? try arc, the ai2 reasoning challenge},
  author={Clark, Peter and Cowhey, Isaac and Etzioni, Oren and Khot, Tushar and Sabharwal, Ashish and Schoenick, Carissa and Tafjord, Oyvind},
  journal={arXiv preprint arXiv:1803.05457},
  year={2018}
}

\bibliographystyle{configuration/iclr2027_conference}
\newpage
\appendix
\section{Limitations}
\label{app:limitations}

Our experiments cover models up to 9B parameters, leaving the effectiveness
of \method at substantially larger scales unverified. Although we evaluate
mathematical reasoning, Countdown, and coding, broader task coverage is needed
to assess its generality beyond these verifiable domains. Our critic evaluation
also reflects a trade-off under computational constraints: we sample MC
continuations at selected anchors on a subset of prompts, balancing trajectory
coverage against the accuracy of each value estimate. These diagnostics capture
value changes between anchors but can miss fluctuations within each interval,
and finite MC samples introduce estimation noise. Finally, the current
formulation assumes binary terminal rewards and $\gamma=1$; extending it to
dense or non-binary rewards requires further investigation.

\section{Value Functions and Credit Assignment}
\label{app:value_credit}

\subsection{What Does the Value Function Learn?}
\label{app:value_derivation}

Consider the population value-regression objective under the trajectory
distribution induced by policy $\pi$:
\begin{equation}
    \mathcal{L}(V)
    =
    \mathbb{E}_{\tau\sim\pi}
    \left[
        \bigl(V(\ss_t)-G_t\bigr)^2
    \right].
\end{equation}
Conditioned on a particular state $\ss_t=\ss$, its contribution is
\begin{equation}
    \mathcal{L}_{\ss}(V)
    =
    \mathbb{E}_{\pi}
    \left[
        \bigl(V(\ss)-G_t\bigr)^2
        \mid \ss_t=\ss
    \right].
\end{equation}
Differentiating with respect to $V(\ss)$ gives
\begin{equation}
    \frac{\partial \mathcal{L}_{\ss}}{\partial V(\ss)}
    =
    2\left(
        V(\ss)-\mathbb{E}_{\pi}[G_t\mid \ss_t=\ss]
    \right).
\end{equation}
Therefore, the population minimizer is
\begin{equation}
    V^*(\ss_t)
    =
    \mathbb{E}_{\pi}[G_t\mid \ss_t]
    =
    V^\pi(\ss_t).
\end{equation}

For a terminal-only reward, $r_k=0$ for $k<T$ and $r_T=R(\yy)$, so
\begin{equation}
    G_t=\gamma^{T-t}R(\yy),
\end{equation}
and hence
\begin{equation}
    V^\pi(\ss_t)
    =
    \mathbb{E}_{\yy_{\ge t}\sim\pi(\cdot\mid \ss_t)}
    \left[
        \gamma^{T-t}R(\yy)
    \right].
\end{equation}
When $\gamma=1$ and $R(\yy)\in\{0,1\}$, this reduces to
\begin{align}
    V^\pi(\ss_t)
    &=
    \mathbb{E}_{\pi}[R(\yy)\mid \ss_t] \\
    &=
    \Pr_{\pi}\!\left(R(\yy)=1\mid \ss_t\right).
\end{align}
Thus, the critic estimates the success probability of continuations produced
by $\pi$.
In PPO, the relevant policy is typically the rollout policy $\pi_{\mathrm{old}}$, so the value target changes as the policy is updated.

\subsection{Value Changes and PPO Credit Assignment}
\label{app:value_change_credit}

We connect changes in prefix values to token-level advantages and to the
GAE signals used by PPO. For autoregressive generation, sampling token
$a_t$ deterministically produces the next state
$\ss_{t+1}=(\ss_t,a_t)$. For nonterminal steps ($t<T$), zero
intermediate rewards and $\gamma=1$ give
\begin{align}
    Q^\pi(\ss_t,a_t)
    &= \EEb{\pi}{G_t \mid \ss_t,a_t}
     = \EEb{\pi}{G_{t+1} \mid \ss_{t+1}}
     = V^\pi(\ss_{t+1}),
\end{align}
so
\begin{equation}
    A^\pi(\ss_t,a_t)
    =V^\pi(\ss_{t+1})-V^\pi(\ss_t)
    \doteq\Delta V_t^\pi.
    \label{eq:app_true_advantage}
\end{equation}
Thus, the true token-level advantage is the change in the probability of
eventual success induced by the token.

Let $\lambda_{\mathrm{actor}}$ denote the GAE coefficient used for the
policy update, and let $\widehat V$ be a fixed critic, with
$\widehat V_t=\widehat V(\ss_t)$. With the value after
termination set to zero,
\begin{equation}
    \widehat A_t^{(\lambda_{\mathrm{actor}})}
    =\sum_{\ell=0}^{T-t}
      \lambda_{\mathrm{actor}}^\ell\widehat\delta_{t+\ell},
    \qquad
    \widehat\delta_k=r_k+\widehat V_{k+1}-\widehat V_k.
    \label{eq:app_general_actor_gae}
\end{equation}
For terminal reward $R$, this can be written directly in terms of local value
changes:
\begin{equation}
\begin{aligned}
    \widehat A_t^{(\lambda_{\mathrm{actor}})}={}&
    \sum_{\ell=0}^{T-t-1}\lambda_{\mathrm{actor}}^\ell
    \bigl(\widehat V_{t+\ell+1}-\widehat V_{t+\ell}\bigr)\\
    &+\lambda_{\mathrm{actor}}^{T-t}
    \bigl(R-\widehat V_T\bigr).
\end{aligned}
\label{eq:app_general_value_change_gae}
\end{equation}
\paragraph{How $\lambda<1$ affects credit assignment.}
At $\lambda_{\mathrm{actor}}=0$, the nonterminal signal is the current
local change,
\begin{equation}
    \widehat A_t^{\mathrm{GAE}}(0)
    =\widehat V(\ss_{t+1})-\widehat V(\ss_t)
    =\Delta\widehat V_t,
    \label{eq:advantage_value_change}
\end{equation}
whereas at $\lambda_{\mathrm{actor}}=1$ the terms telescope to
$R-\widehat V(\ss_t)$. More generally, intermediate values of
$\lambda_{\mathrm{actor}}$ geometrically weight local changes at increasing
distances. At the terminal step, all settings give
$\widehat A_T=R-\widehat V_T$, since the value after termination is zero.
With exact values, the nonterminal one-step estimator equals the true
advantage in Equation~\ref{eq:app_true_advantage}.

For $0<\lambda_{\mathrm{actor}}<1$, a value change $\ell$ steps ahead
contributes to the current token's advantage with weight
$\lambda_{\mathrm{actor}}^\ell$. Smaller $\lambda_{\mathrm{actor}}$ therefore
places more emphasis on nearby changes and attenuates the distant terminal
signal. Since advantages weight token gradients in
Equation~\ref{eq:policy_gradient}, errors in these nearby changes can distort
the direction or strength of the policy update. Accurately tracking value
evolution can improve these credit signals, especially when the terminal
outcome is many steps away.

This also changes the relative credit between adjacent tokens. For $t<T$,
the GAE recursion gives
\begin{equation}
\begin{aligned}
    \widehat A_t^{(\lambda_{\mathrm{actor}})}
    -\widehat A_{t+1}^{(\lambda_{\mathrm{actor}})}
    ={}&\Delta\widehat V_t
    -(1-\lambda_{\mathrm{actor}})
    \widehat A_{t+1}^{(\lambda_{\mathrm{actor}})}.
\end{aligned}
\label{eq:app_adjacent_credit_general_lambda}
\end{equation}
Thus, for $\lambda_{\mathrm{actor}}<1$, adjacent-token credit depends on both
the local value change and the downstream advantage; the local change alone
determines their difference only at $\lambda_{\mathrm{actor}}=1$.

\paragraph{The $\lambda=1$ case.}
For completeness, when $\gamma=\lambda=1$ as in the main experiments,
\begin{equation}
    \widehat A_t
    =R(\yy)-\widehat V_t,
    \label{eq:app_gae_terminal}
\end{equation}
and, if $\widehat V=V^\pi$, then for $t<T$
\begin{align}
    \EEb{\pi}{\widehat A_t\mid\ss_t,a_t}
    &=V^\pi(\ss_{t+1})-V^\pi(\ss_t)
    =A^\pi(\ss_t,a_t).
    \label{eq:app_expected_gae}
\end{align}
The sampled $\lambda=1$ estimator therefore recovers the token-level value
change in expectation. For adjacent positions in the same response, the
shared terminal return also gives the sample-wise ordering identity
\begin{equation}
    \widehat A_t-\widehat A_{t+1}
    =\widehat V_{t+1}-\widehat V_t
    =\Delta\widehat V_t.
    \label{eq:app_advantage_difference}
\end{equation}

\subsection{Monte Carlo Evaluation}
\label{app:coarse_value_change}

To visualize how values evolve over the course of reasoning, we evaluate
states at a fixed set of relative positions along each response.
For a response containing $T$ tokens, let
\[
    0 < \rho_1 < \rho_2 < \cdots < \rho_M \leq 1
\]
denote predefined fractions of the response length.
The corresponding token positions are
\begin{equation}
    t_j = \left\lfloor \rho_j T \right\rfloor,
    \qquad j=1,\ldots,M.
    \label{eq:eval_position}
\end{equation}
Thus, responses of different lengths are evaluated at comparable stages of
generation.

For two consecutive evaluation points, we define the coarse-grained value
change as
\begin{equation}
    \Delta^{(\rho)}_j V
    \doteq
    V(\ss_{t_{j+1}})
    -
    V(\ss_{t_j}).
    \label{eq:coarse_value_change}
\end{equation}

Although Equation~\ref{eq:coarse_value_change} measures the change over a
segment of the response rather than between two adjacent tokens, it is
directly related to the token-level value changes discussed in
Section~\ref{sec:characterizing_failure}.
By telescoping,
\begin{equation}
    \Delta^{(\rho)}_j V
    =
    \sum_{t=t_j}^{t_{j+1}-1}
    \left[
        V(\ss_{t+1})-V(\ss_t)
    \right].
    \label{eq:coarse_value_telescoping}
\end{equation}
Hence, the change between two evaluation points summarizes the cumulative
token-level value evolution over the corresponding segment of reasoning.

\paragraph{Monte Carlo value estimates.}
For each selected state $\ss_{t_j}$, we estimate its policy value by sampling
$K_j$ valid continuations from the rollout policy at anchor $j$:
\begin{equation}
    \widehat V_{\mathrm{MC}}(\ss_{t_j})
    \doteq
    \frac{1}{K_j}
    \sum_{k=1}^{K_j}
    R\!\left(\yy^{(k)}\right),
    \qquad
    \yy^{(k)} \sim \pi(\cdot \mid \ss_{t_j}).
    \label{eq:mc_value_at_fraction}
\end{equation}
Because
\begin{equation}
    V^\pi(\ss_{t_j})
    =
    \EEb{\pi}{R(\yy)\mid S_{t_j}=\ss_{t_j}},
\end{equation}
$\widehat V_{\mathrm{MC}}(\ss_{t_j})$ is a Monte Carlo estimate of the same
state value targeted by the PPO critic.

We then compute the Monte Carlo value change between consecutive evaluation
points as
\begin{equation}
    \Delta^{(\rho)}_j V_{\mathrm{MC}}
    \doteq
    \widehat V_{\mathrm{MC}}(\ss_{t_{j+1}})
    -
    \widehat V_{\mathrm{MC}}(\ss_{t_j}),
    \label{eq:mc_coarse_value_change}
\end{equation}
and the corresponding PPO-predicted change as
\begin{equation}
    \Delta^{(\rho)}_j V_{\mathrm{PPO}}
    \doteq
    V_{\phi}(\ss_{t_{j+1}})
    -
    V_{\phi}(\ss_{t_j}).
    \label{eq:ppo_coarse_value_change}
\end{equation}

Figure~\ref{fig:illustration}(a) illustrates these estimates for a correct
and an incorrect response.

\paragraph{Connection to GAE advantage ordering.}
Under terminal-only rewards and $\gamma=\lambda=1$, let
$\widehat A_t^{\mathrm{GAE}}[V]=R(\yy)-V(\ss_t)$ denote the advantage
computed using a value function $V$, with zero terminal-state value.
Applying Equation~\ref{eq:app_advantage_difference} across the
interval between two anchors gives
\begin{equation}
    \widehat A_{t_j}^{\mathrm{GAE}}[V]
    -\widehat A_{t_{j+1}}^{\mathrm{GAE}}[V]
    =V(\ss_{t_{j+1}})-V(\ss_{t_j})
    =\Delta_j^{(\rho)}V.
    \label{eq:anchor_advantage_difference}
\end{equation}
The endpoint identity also holds when MC estimates replace the critic values
at the two anchors. Thus, when the critic and MC changes are both nonzero and
have opposite signs at the same anchor pair for the same rollout policy,
the critic-based advantages reverse the ordering given by the MC-based
reference. This reference uses sampled estimates rather than exact policy
values. The comparison concerns relative advantages at the anchors; it does
not establish the sign of every individual token-level value change within
the interval.

\section{Bayesian Value Evolution and TPPO Details}
\label{app:tppo_details}

\subsection{Proof of the Bayesian Prefix-Value Update}
\label{app:bayesian_value_update}

Because the next state is the deterministic concatenation
$\ss_{t+1}=(\ss_t,a_t)$,
\begin{align}
    V^{\pi_\theta}(\ss_{t+1})
    &=\PPb{\pi_\theta}{R=1\mid\ss_t,a_t} \\
    &=\frac{\PPb{\pi_\theta}{R=1,a_t\mid\ss_t}}{\PPb{\pi_\theta}{a_t\mid\ss_t}},
\end{align}
where the denominator is positive by assumption. If
$0<V^{\pi_\theta}(\ss_t)<1$, the numerator and denominator factor as
\begin{align}
    \PPb{\pi_\theta}{R=1,a_t\mid\ss_t}
    &=V^{\pi_\theta}(\ss_t)\PPb{\pi_\theta}{a_t\mid\ss_t,R=1},\\
    \PPb{\pi_\theta}{a_t\mid\ss_t}
    &=V^{\pi_\theta}(\ss_t)\PPb{\pi_\theta}{a_t\mid\ss_t,R=1}\notag\\
    &\quad+\bigl(1-V^{\pi_\theta}(\ss_t)\bigr)
       \PPb{\pi_\theta}{a_t\mid\ss_t,R=0},
\end{align}
which proves Equation~\ref{eq:bayesian_value_update}. If the current value is
zero, every positive-probability action has zero joint success probability;
the value-one case follows analogously from zero joint failure probability.

Under the positivity assumptions in Proposition~\ref{prop:logit_value_update},
let $\Lambda_t=\PPb{\pi_\theta}{a_t\mid\ss_t,R=1}/
\PPb{\pi_\theta}{a_t\mid\ss_t,R=0}$. Dividing the posterior success
probability by its complement gives
\begin{equation}
    \frac{V^{\pi_\theta}(\ss_{t+1})}{1-V^{\pi_\theta}(\ss_{t+1})}
    =
    \frac{V^{\pi_\theta}(\ss_t)}{1-V^{\pi_\theta}(\ss_t)}\Lambda_t.
\end{equation}
Taking logarithms proves Equation~\ref{eq:logit_value_update}; iterating this identity
gives the cumulative decomposition in Corollary~\ref{cor:value_decomposition}.

\subsection{Learning Signals in the Cumulative Parameterization}
\label{app:implicit_delta_learning}

\method learns position-wise scalar outputs through supervision on the
reconstructed values. Here we derive the suffix-aggregated learning signal
for each logit increment. This derivation explains how terminal outcomes
train the cumulative parameterization; it does not establish recovery of
the true Bayesian log-likelihood ratios.

\paragraph{Learning signal for each increment.}
For a response of length $T$, let
\begin{equation}
    \widehat z_t
    =
    \widehat z_1
    +
    \sum_{j=2}^{t}\widehat\Delta_j,
    \qquad
    \widehat V_t=\sigma(\widehat z_t).
\end{equation}
Under a terminal binary outcome $R\in\{0,1\}$, the BCE objective is
\begin{equation}
    \mathcal L_V
    =
    \frac{1}{T}
    \sum_{t=1}^{T}
    \ell(\widehat z_t,R),
\end{equation}
where
\begin{equation}
    \ell(\widehat z_t,R)
    =
    -R\log\sigma(\widehat z_t)
    -(1-R)\log\!\left(1-\sigma(\widehat z_t)\right).
\end{equation}

For $i\geq2$, the increment $\widehat\Delta_i$ affects every
$\widehat z_t$ with $t\geq i$, and
\begin{equation}
    \frac{\partial \widehat z_t}
         {\partial \widehat\Delta_i}
    =
    \mathbbm{1}\{t\geq i\}.
\end{equation}
Therefore, by the chain rule,
\begin{equation}
\begin{aligned}
    \frac{\partial \mathcal L_V}
         {\partial \widehat\Delta_i}
    &=
    \frac{1}{T}
    \sum_{t=1}^{T}
    \frac{\partial \ell(\widehat z_t,R)}
         {\partial \widehat z_t}
    \frac{\partial \widehat z_t}
         {\partial \widehat\Delta_i} \\
    &=
    \frac{1}{T}
    \sum_{t=i}^{T}
    \frac{\partial \ell(\widehat z_t,R)}
         {\partial \widehat z_t}.
\end{aligned}
\label{eq:implicit_delta_gradient}
\end{equation}
For BCE with logits,
\begin{equation}
    \frac{\partial \ell(\widehat z_t,R)}
         {\partial \widehat z_t}
    =
    \sigma(\widehat z_t)-R
    =
    \widehat V_t-R,
\end{equation}
which gives
\begin{equation}
    \frac{\partial \mathcal L_V}
         {\partial \widehat\Delta_i}
    =
    \frac{1}{T}
    \sum_{t=i}^{T}
    \left(\widehat V_t-R\right).
    \label{eq:implicit_delta_gradient_bce}
\end{equation}

Hence, each increment receives the aggregate value-learning signal from
the suffix of reconstructed values that depends on it.
No direct target for $\widehat\Delta_i$ is required.
Although all positions in a response share the same terminal outcome $R$,
different increments receive different gradients because they affect
different suffixes of the value sequence.

Indeed, adjacent increments satisfy
\begin{equation}
\begin{aligned}
    \frac{\partial \mathcal L_V}
         {\partial \widehat\Delta_i}
    -
    \frac{\partial \mathcal L_V}
         {\partial \widehat\Delta_{i+1}}
    &=
    \frac{1}{T}
    \left(\widehat V_i-R\right).
\end{aligned}
\label{eq:adjacent_delta_gradient}
\end{equation}
Thus, the cumulative parameterization converts terminal value supervision
into position-dependent, suffix-aggregated learning signals for the
token-wise increments.

\section{Experimental Details and Additional Results}
\label{app:experimental_details}

\subsection{Training Setup}
\label{app:training_setup}

\paragraph{Training data and rewards.}
Mathematical reasoning uses Polaris with binary answer-verification rewards.
Scientific coding uses Qwen3.5-9B on a multi-turn dataset of 8,178 main
problems and 37,820 subproblems, with binary code-execution rewards and a
120-second training timeout. Thinking is enabled and the training seed is 42.
\begin{table}[!htbp]
    \centering
    \caption{\textbf{Training, rollout, and optimization settings.}
    Shared settings apply to both domains unless a domain-specific value is shown.}
    \label{tab:training_hyperparameters}
    \small
    \setlength{\tabcolsep}{6pt}
    \renewcommand{\arraystretch}{1.15}
    \begin{tabular}{@{}>{\raggedright\arraybackslash}p{0.26\linewidth}
                    >{\raggedright\arraybackslash}p{0.39\linewidth}
                    >{\raggedright\arraybackslash}p{0.27\linewidth}@{}}
        \toprule
        \rowcolor{tableheader}
        \textbf{Parameter} & \textbf{Math} & \textbf{Scientific coding} \\
        \midrule
        Maximum response length & 8,192 & 20,000 \\
        Temperature & 1.0 & 1.0 \\
        Top-$p$ & 0.95 & 1.0 \\
        Batch size & 64 prompts (Base) / 128 prompts (4B) & 32 main problems $\times$ 8 trajectories \\
        Minibatch size & 32 prompts (Base) / 64 prompts (4B) & 256 responses \\
        \addlinespace[3pt]
        Actor learning rate & $10^{-6}$ & $10^{-6}$ \\
        Critic learning rate & $4\times10^{-6}$ & $10^{-5}$ \\
        Learning-rate schedule & Constant & Constant \\
        Weight decay & 0.1 & 0.1 \\
        Gradient clipping & 1.0 & 1.0 \\
        KL / entropy coefficient & $0$ / $0$ & $0$ / $0$ \\
        Importance-weight upper threshold & 2.0 & 2.0 \\
        \addlinespace[3pt]
        Training framework & Slime / Megatron & VERL\footnotemark / Megatron \\
        GPUs per node & 8 & 8 \\
        \bottomrule
    \end{tabular}
\end{table}
\footnotetext{https://github.com/volcengine/verl}

\paragraph{Baselines.}
PPO uses token-level values, GAE with $\gamma=\lambda=1$, clipped MSE, and
clips the actor's policy probability ratio
$\rho_t(\theta)=\pi_\theta(a_t\mid \ss_t)/\pi_{\mathrm{old}}(a_t\mid \ss_t)$ in its
surrogate objective to $[1-0.2,1+0.2]$. VAPO uses the Clip-Higher variant,
which clips this same policy ratio asymmetrically with
$\epsilon_{\mathrm{low}}=0.2$ and $\epsilon_{\mathrm{high}}=0.28$, together
with adaptive actor $\lambda(L)$ and an NLL term on correct responses. GRPO standardizes
rewards within response groups, while SPPO broadcasts a prompt-level success
baseline to all response tokens.

\subsection{Additional Results}
\label{app:additional_results}

In Tables~\ref{tab:main_results} and~\ref{tab:main_results_n1}, Avg. averages
the seven in-domain benchmarks equally, counting each AIME year separately;
OOD Avg. averages the three OOD benchmarks equally.

\begin{figure}[h]
    \centering
    \includegraphics[width=0.5\linewidth]{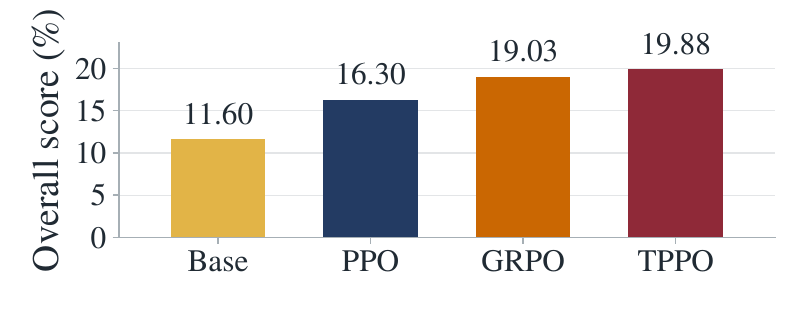}
    \caption{\small
        \textbf{Additional results on Llama3.2-3B-Instruct.}
        Overall scores for the base model, PPO, GRPO, and \method (\%).
    }
    \label{fig:llama32_results}
\end{figure}

Figure~\ref{fig:llama32_results} reports the additional Llama3.2-3B-Instruct
comparison on mathematical reasoning, complementing the model comparisons in
Section~\ref{sec:main_results}. The overall scores are 11.60 for the base
model, 16.30 for PPO, 19.03 for GRPO, and 19.88 for \method; thus, \method
improves over PPO by 3.58 points on this model.

\begin{figure}[h]
    \centering
    \includegraphics[width=0.7\linewidth]{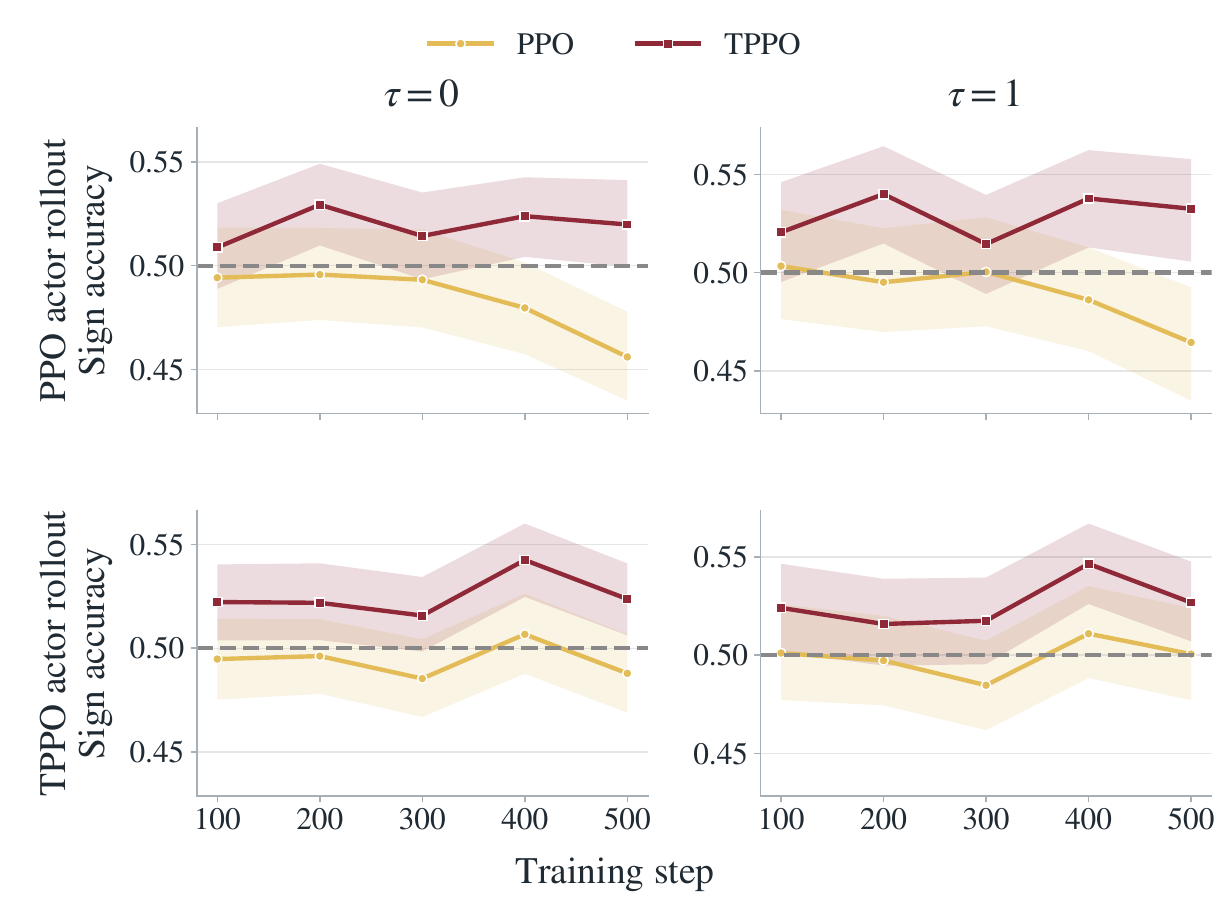}
    \caption{\small
        \textbf{Value-change sign accuracy by rollout source.}
        Rows distinguish PPO-actor and \method-actor rollouts; columns show
        the SNR thresholds $\tau=0$ and $\tau=1$.
    }
    \label{fig:rollout_source_sign_accuracy}
\end{figure}

\paragraph{Controlling for rollout source.}
Figure~\ref{fig:rollout_source_sign_accuracy} compares PPO and \method
critics on the same actor rollouts. This additional result controls for the
fact that the main comparison evaluates each critic on responses generated by
its own actor.

\subsection{Diagnostic Metrics}
\label{app:diagnostic_metrics}

\paragraph{Value-estimation metrics on test responses.}
\label{app:value_quality_metrics}
In Figure~\ref{fig:value_estimates_test}, value accuracy is the fraction of
prefixes satisfying $|\widehat V-\widehat V_{\mathrm{MC}}|\leq0.05$ at each
evaluated reasoning position.

\paragraph{Sampling protocol.}
The diagnostics use Qwen3-8B-Base actor updates at steps 100, 200, 300,
400, and 500. At each step, the 64 prompts are a fixed subset of the
training set, and exactly the same subset is used for PPO and \method at
all five steps. Each method's own actor generates
eight responses per prompt, giving 512 responses per method and checkpoint.
Each response has nine MC anchors at approximately 10\%, 20\%, $\ldots$,
90\% of its length. This yields 4,608 anchors and 4,096 adjacent-anchor
intervals per method and checkpoint. Figure~\ref{fig:illustration}(a)
shows individual examples; the aggregate diagnostics use this sampling protocol.

Each anchor is allocated a budget of 128 MC continuations under the
corresponding actor. Across both methods and all five checkpoints, 46,080 anchors have 128 valid samples.
The total number of valid MC samples is $5.90\mathrm{M}$. MC means and standard errors use the
actual valid sample count at each anchor.

These training-set diagnostics measure value-change sign agreement and
agreement with MC-based reference advantages, as defined below.
Figure~\ref{fig:rollout_source_sign_accuracy} additionally controls for rollout
source when evaluating value-change sign agreement.

\paragraph{SNR-conditioned value-change sign accuracy.}
Let $\widehat V_{\mathrm{MC},j}$ be the MC value estimate at anchor $j$,
and let $s_j$ and $n_j$ denote the standard deviation of its sampled
continuation returns and the number of MC samples, respectively.
For adjacent anchors, define
\begin{equation}
\begin{aligned}
    \Delta\mathrm{MC}_j
    &= \widehat V_{\mathrm{MC},j+1}-\widehat V_{\mathrm{MC},j}, \\
    \mathrm{SNR}_j
    &= \frac{|\Delta\mathrm{MC}_j|}
    {\sqrt{s_j^2/n_j+s_{j+1}^2/n_{j+1}}}.
\end{aligned}
\label{eq:mc_change_snr}
\end{equation}
The denominator estimates the standard error of the MC difference;
the calculation omits the covariance between the two anchor estimates.
For a critic-predicted change $\Delta V_j$, we compute the empirical
conditional sign agreement
\begin{equation}
    \operatorname{SignAcc}(\tau)
    = \Pr\!\left[
        \operatorname{sign}(\Delta V_j)
        =\operatorname{sign}(\Delta\mathrm{MC}_j)
        \,\middle|\, \mathrm{SNR}_j>\tau
    \right].
\label{eq:snr_conditioned_sign_accuracy}
\end{equation}
The threshold $\tau=0$ retains nonzero MC changes, while $\tau=1$
retains changes whose magnitude exceeds one estimated standard error.
The selection depends solely on the MC samples and does not use critic
predictions. Figure~\ref{fig:value_changes_metrics} reports each method's
absolute sign accuracy under these two thresholds. Filtering is performed
separately for each method, and each accuracy is normalized by its own
retained interval count, rather than by all 4,096 candidates.
Table~\ref{tab:diagnostic_interval_counts} lists these denominators.

\begin{table}[!ht]
    \centering
    \small
    \caption{\textbf{Intervals retained by SNR filtering.} Counts per method
    and checkpoint, with percentages of the 4,096 candidate intervals
    in parentheses.}
    \label{tab:diagnostic_interval_counts}
    \setlength{\tabcolsep}{8pt}
    \begin{tabular}{@{}rcccc@{}}
        \toprule
        & \multicolumn{2}{c}{$\tau=0$} & \multicolumn{2}{c}{$\tau=1$} \\
        \cmidrule(lr){2-3}\cmidrule(l){4-5}
        Actor update & PPO & \method & PPO & \method \\
        \midrule
        100 & 2,765 (67.5\%) & 2,843 (69.4\%) & 1,655 (40.4\%) & 1,972 (48.1\%) \\
        200 & 2,403 (58.7\%) & 2,968 (72.5\%) & 1,586 (38.7\%) & 1,930 (47.1\%) \\
        300 & 2,186 (53.4\%) & 2,793 (68.2\%) & 1,468 (35.8\%) & 1,860 (45.4\%) \\
        400 & 1,944 (47.5\%) & 2,755 (67.3\%) & 1,323 (32.3\%) & 1,852 (45.2\%) \\
        500 & 1,991 (48.6\%) & 2,767 (67.6\%) & 1,339 (32.7\%) & 1,830 (44.7\%) \\
        \bottomrule
    \end{tabular}
\end{table}

\paragraph{MC-based reference advantages.}
For these diagnostics, $\gamma=\lambda=1$, intermediate rewards are zero,
there is no KL reward shaping, and the value after termination is zero.
GAE therefore reduces to the terminal outcome minus the value at the anchor:
\begin{equation}
    \widehat A_{t_j}=R-\widehat V_{t_j},
    \qquad
    \widehat A_{t_j}^{\mathrm{MC}}=R-\widehat V_{\mathrm{MC}}(\ss_{t_j}).
    \label{eq:mc_reference_advantage}
\end{equation}
Here $R$ is the binary terminal reward of the original response.
The reference replaces the critic value with the MC estimate at the same
anchor. Both metrics use all 4,608 anchors per method and checkpoint.
For cosine similarity, we concatenate the critic advantages and reference
advantages into two vectors and compute their overall cosine similarity.
Sign accuracy is the fraction of anchors at which their signs agree;
zero values are included with $\operatorname{sign}(0)=0$.
Figure~\ref{fig:advantage_metrics_training} reports each method's absolute
metrics, while Table~\ref{tab:diagnostic_gains} reports their differences.

\paragraph{Bootstrap estimation of metric gains.}
To quantify uncertainty in the gains over PPO, we draw 2,000 bootstrap
resamples of the 512 response indices with replacement, using random seed
1234. Each sampled response contributes all its anchors and intervals,
preserving their within-response dependence. The same resampled
\texttt{sample\_index} sequence is used for both methods; each method
retains its own generated responses, MC estimates, and SNR-filtered
intervals. We recompute the aggregate metrics in each resample and take
the 2.5th and 97.5th percentiles of the \method-minus-PPO differences.
These intervals describe response-resampling uncertainty conditional on
the fixed prompts and trained checkpoints; the resampling unit is a
response, with eight responses per prompt.

Across actor update steps 200--500, the 95\% intervals for advantage sign-accuracy
gains and value-change sign-accuracy gains under both SNR thresholds
are above zero. These are intervals for metric differences, distinct from
the individual-method curves in Figures~\ref{fig:value_changes_metrics}
and~\ref{fig:advantage_metrics_training}.

\begin{table}[!ht]
    \centering
    \small
    \caption{\textbf{Diagnostic gains of \method over PPO.}
    Sign-accuracy gains are in percentage points; cosine gains are
    absolute differences. Rows correspond to actor update steps.}
    \label{tab:diagnostic_gains}
    \setlength{\tabcolsep}{9pt}
    \begin{tabular}{@{}rcccc@{}}
        \toprule
        & \multicolumn{2}{c}{Advantage} & \multicolumn{2}{c}{Value-change sign accuracy} \\
        \cmidrule(lr){2-3}\cmidrule(l){4-5}
        Actor update & Cosine & Sign accuracy & $\tau=0$ & $\tau=1$ \\
        \midrule
        100 & $-0.0228$ & $+1.76$ & $-0.12$ & $+0.06$ \\
        200 & $+0.1189$ & $+15.34$ & $+3.90$ & $+4.94$ \\
        300 & $+0.0574$ & $+16.60$ & $+6.35$ & $+7.40$ \\
        400 & $+0.1089$ & $+19.92$ & $+4.62$ & $+5.29$ \\
        500 & $+0.0485$ & $+19.44$ & $+6.76$ & $+6.23$ \\
        \bottomrule
    \end{tabular}

\end{table}

\section{Additional Ablations and Discussion}
\label{app:additional_discussion}

\subsection{Shared and Separate Output Heads}
\label{app:shared_head}

Our default critic uses the shared scalar head described in the experimental
setup (Section~\ref{sec:experimental_setup}): the first output supplies the
initial logit and subsequent outputs supply logit increments.

\paragraph{Separate-head parameterization.}
We also tested a variant with two independent scalar heads,
\begin{equation}
    \widehat z_1 = f_{\phi_0}(\hh_1),
    \qquad
    \widehat\Delta_t = f_{\phi_\Delta}(\hh_t),
    \quad t\geq2.
\end{equation}
This variant uses separate output parameters for the initial logit and
subsequent increments. The cumulative reconstruction and terminal-outcome
supervision remain the same.

\begin{figure}[h]
    \centering
    \includegraphics[width=1.0\linewidth]{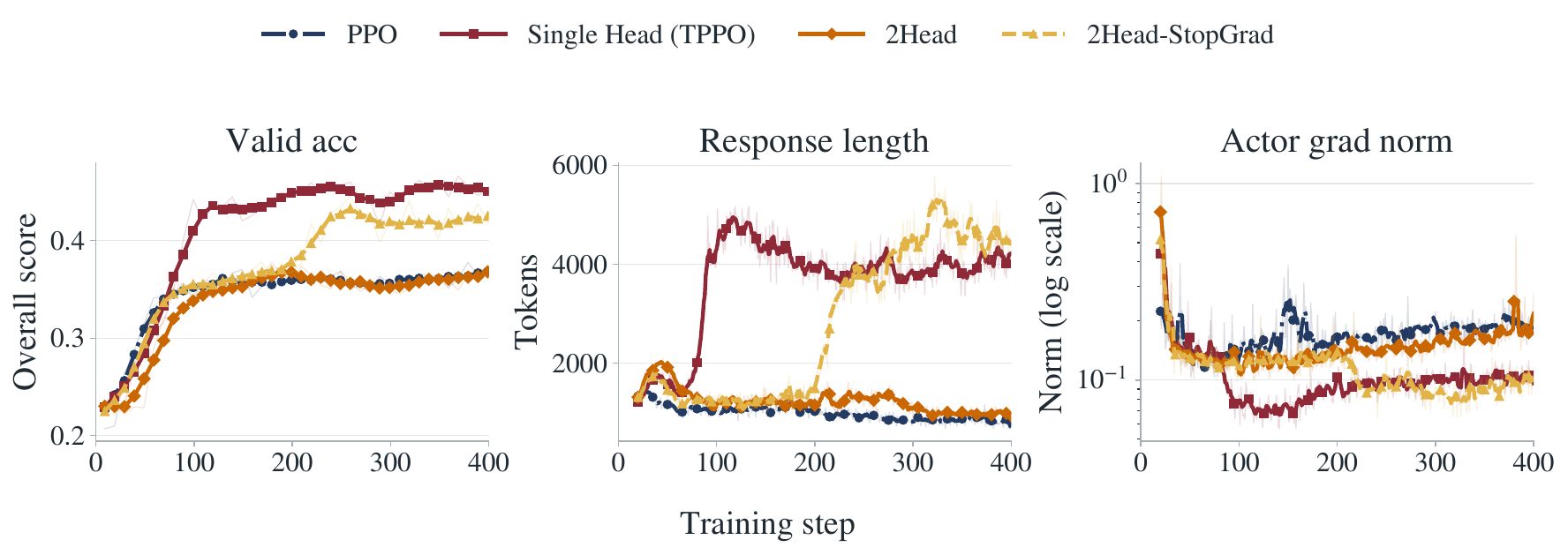}
    \caption{\small
        \textbf{Shared and separate output heads.}
        Validation accuracy, response length, and actor gradient norm
        for PPO, the shared-head TPPO critic, and the two separate-head
        variants shown in the legend.
    }
    \vspace{-1em}
    \label{fig:head_comparison}
\end{figure}

Across runs, the separate-head variant was less stable and learned
different allocations between the initial logit and accumulated increments.
Table~\ref{tab:head_parameterization} summarizes the output statistics:
the initial logits have substantially larger magnitude and variance,
while the increment scales remain comparable.
Figure~\ref{fig:head_comparison} compares validation accuracy, response
length, and actor gradient norms for the head variants and PPO.

These observations motivate using a shared head in all main experiments.
Sharing ties the initial-logit and increment outputs through the same
parameters, without additional losses or regularization terms. The Bayesian
identity itself does not prescribe whether the heads should be shared.

\begin{table}[h]
\centering
\small
\setlength{\tabcolsep}{8pt}
\renewcommand{\arraystretch}{1.15}
\caption{\textbf{Output statistics for shared and separate heads.}
The separate head has substantially larger initial-logit magnitude and
variance, while increment statistics remain comparable.}
\label{tab:head_parameterization}
\begin{tabular}{@{}lrrrr@{}}
\toprule
& \multicolumn{2}{c}{Initial logit} & \multicolumn{2}{c}{Increment} \\
\cmidrule(lr){2-3}\cmidrule(l){4-5}
Head & Mean & SD & Mean & SD \\
\midrule
\rowcolor{methodlight}
Shared & $-0.0273$ & $0.0052$ & $-0.00054$ & $0.00210$ \\
Separate & $-0.8657$ & $0.3016$ & $-0.00132$ & $0.00257$ \\
\bottomrule
\end{tabular}
\end{table}

\subsection{Actor GAE Coefficients}
\label{app:actor_gae_settings}

For the comparison in Section~\ref{sec:discussion} and
Figure~\ref{fig:ablate_lambda}(b), we evaluate
$\lambda_{\mathrm{actor}}\in\{1,0.95,0\}$ while keeping the critic's
$\lambda$ fixed at 1.
The critic is warmed up for 200 steps with the actor held fixed before
joint actor--critic training begins.
The $\lambda_{\mathrm{actor}}=0$ condition computes advantages at the
segment level: with zero intermediate rewards and $\gamma=1$, each
nonterminal segment TD residual is the predicted value difference between
its boundaries. Following the segment-level credit assignment used in
VinePPO~\citep{kazemnejad2024vineppo}, we assign the same segment advantage to
every token within that segment. The other settings use the corresponding GAE construction
in Appendix~\ref{app:value_change_credit}.

\end{document}